\documentclass[twoside,11pt]{article}

\usepackage[abbrvbib, preprint]{jmlr2e}

\usepackage[utf8]{inputenc} % allow utf-8 input
\usepackage[T1]{fontenc}    % use 8-bit T1 fonts
\usepackage{url}            % simple URL typesetting
\usepackage{booktabs}       % professional-quality tables
\usepackage{amsfonts}       % blackboard math symbols
\usepackage{nicefrac}       % compact symbols for 1/2, etc.
\usepackage{microtype}      % microtypography

\usepackage{enumitem}
\usepackage{adjustbox}
\usepackage[leftcaption]{sidecap}
\usepackage{color}

\usepackage{times}
\usepackage{graphicx}
\usepackage{float}
\usepackage{wrapfig}
\usepackage{amsmath}
\usepackage{algorithm,algpseudocode}
\usepackage{tabularx}
\usepackage{bm,xspace}
\usepackage{comment}
\usepackage{verbatim}
\usepackage{multirow}
\usepackage{balance}
\usepackage{url}
\usepackage{booktabs}
\usepackage{etoolbox,siunitx}
\usepackage{calc}
\usepackage{pifont,hologo}
\usepackage{nicefrac}

\usepackage[super]{nth}
\usepackage{nicefrac}
\robustify\bfseries

\usepackage{dsfont}

\newcommand*{\ie}{i.e.\@\xspace}

\def\xx{\mathbf{x}}
\def\yy{\mathbf{y}}

\def\DD{\mathbf{D}}

\def\MM{\mathbf{M}}

\def\SSS{\mathbf{S}}

\def\VV{\mathbf{V}}

\def\Ee{\mathbb{E}}

\def\lL{\mathcal{L}}

\def\pP{\mathcal{P}}

\def\Ee{\mathbb{E}}

\def\btheta{{\bm\theta}}

\def\xx{\mathbf{x}}
\def\yy{\mathbf{y}}

\def\DD{\mathbf{D}}

\def\MM{\mathbf{M}}

\def\SSS{\mathbf{S}}

\def\VV{\mathbf{V}}

\def\Ee{\mathbb{E}}

\DeclareMathOperator*{\argmax}{arg\,max}
\DeclareMathOperator*{\argmin}{arg\,min}

\DeclareMathSymbol{@}{\mathord}{letters}{"3B}

\def\latex/{\LaTeX}
\def\bibtex/{\hologo{BibTeX}}

\newsavebox{\mybox}\newsavebox{\mysim}
\newcommand{\distras}[1]{%
  \savebox{\mybox}{\hbox{\kern3pt$\scriptstyle#1$\kern3pt}}%
  \savebox{\mysim}{\hbox{$\sim$}}%
  \mathbin{\overset{#1}{\kern\z@\resizebox{\wd\mybox}{\ht\mysim}{$\sim$}}}%
}

\newcommand{\ourdata}{\textsc{Firehose}\xspace}

\newcommand{\ourproblem}{\textsc{POLL}\xspace}
\newcommand{\ourmethod}{\textsc{ConGraD}\xspace}

\jmlrheading{1}{2000}{1-48}{4/00}{11/20}{firehose}{Hu and Sener and Sha and Koltun}

\ShortHeadings{Drinking from a Firehose}{Hu and Sener and Sha and Koltun}
\firstpageno{1}

\begin{document}

\title{Drinking from a Firehose: Continual Learning with Web-scale Natural Language}

\author{\name Hexiang Hu \email hexiangh@usc.edu \\
\addr University of Southern California
\AND
\name Ozan Sener \email ozan.sener@intel.com \\
\addr Intel Labs
\AND
\name Fei Sha \email feisha@usc.edu \\
\addr University of Southern California
\AND
\name Vladlen Koltun \email vladlen.koltun@intel.com \\
\addr Intel Labs
}

\editor{Kevin Murphy and Bernhard Sch{\"o}lkopf}

\maketitle

\begin{abstract}%   <- trailing '%' for backward compatibility of .sty file
Continual learning systems will interact with humans, with each other, and with the physical world through time -- and continue to learn and adapt as they do. An important open problem for continual learning is a large-scale benchmark which enable realistic evaluation of algorithms. In this paper, we study a natural setting for continual learning on a massive scale. We introduce the problem of personalized online language learning (POLL), which involves fitting personalized language models to a population of users that evolves over time. To facilitate research on POLL, we collect massive datasets of Twitter posts. These datasets, Firehose10M and Firehose100M, comprise 100 million tweets, posted by one million users over six years. Enabled by the Firehose datasets, we present a rigorous evaluation of continual learning algorithms on an unprecedented scale. Based on this analysis, we develop a simple algorithm for continual gradient descent (ConGraD) that outperforms prior continual learning methods on the Firehose datasets as well as earlier benchmarks. Collectively, the POLL problem setting, the Firehose datasets, and the ConGraD algorithm enable a complete benchmark for reproducible research on web-scale continual learning.
\end{abstract}


\section{Introduction}
Machine learning is traditionally considered in the disjoint phases of training where learning happens, and testing where decision making happens. Although this framework is fruitful, it is limited for agents which interact with humans and the physical world through time. Physical world, humans, and other artificial agents evolve through time, making continually learning a necessity. The perspective of continuously learning and evolving through time was initially referred to as lifelong learning by \citet{Thrun1}. Following the initial conception, various algorithms and continual learning systems have been developed \citep{LifelongLearningBook}. 

Somewhat parallel to the efforts on lifelong learning, connectionist approaches faced an important problem, the stability-plasticity dilemma \citep{stability_plasticity}. Stability refers to retaining knowledge about the previous time instants, whereas plasticity refers to the ability to acquire new knowledge. The more plastic the networks, the more forgetting happens, and the more stable the networks, the less learning happens. The problem of tackling the stability-plasticity dilemma has recently gained attention \citep{kirkpatrick2017overcoming} under the name ``continual learning''. The literature on continual learning has been summarized in recent surveys by \citet{Parisi2019} and \citet{CLSurvey2}. Although continual learning initially referred to sequentially learning multiple disjoint tasks, its definition has been relaxed through time. We use
the relaxed definition of \citet{CLSurvey2}: ``The General Continual Learning setting considers an infinite stream of training data where at each time step, the system receives a (number of) new sample(s) drawn non i.i.d from a current distribution that could itself experience sudden or gradual changes''. Following this general definition, we regard continual learning and lifelong learning as interchangeable within the scope of our paper. We use the term continual learning in the rest of the paper for the sake of consistency.

Continual learning is typically evaluated on supervised image classification tasks which are transformed into a continual setting using either pixel permutation or class splitting \citep{Parisi2019, CLSurvey2}. For permutation, each task is classifying images under fixed pixel permutations while labels stay the same. For splitting, each task is classification over a small and disjoint subset of classes. Although these tasks are convenient, they have major issues that are extensively discussed in the literature \citep{Parisi2019, CLSurvey2}. For example, \citet{Parisi2019} state ``...while there is a large number of approaches capable of alleviating catastrophic forgetting in highly controlled experimental conditions, lifelong learning has not been tackled for more complex scenarios.''. \citet{CLSurvey2} state that ``A largely ignored characteristic of continual learning is being able to learn from a continuous stream of data without offline training of large batches or separate tasks.''

Rather surprisingly, the aforementioned limitations on existing benchmarks are typically not present in traditional lifelong learning systems. One example of such a system is Never Ending Language Learning (NELL) \citep{NELL}. NELL has been continuously crawling web pages and extracting natural language information for more than a decade. NELL also interacted with humans to validate the learned knowledge. The data, dynamics, and interactions NELL uses are all realistic and on a massive scale. One can ask the question \emph{did we catastrophically forget what we learned as a field?}. To be fair, NELL operates with fundamentally different representations: knowledge bases can grow to extreme scales without forgetting, \ie symbolic AI has no stability-plasticity dilemma. Nevertheless, an interesting question to ask is, can we create a continual learning benchmark which shares key desirable aspects with older lifelong learning setups and is applicable to current language learning systems that are based on deep models.

Neural networks have already been utilized to learn natural language from web-scale unstructured data. For example, \citet{gpt2} and \citet{gpt3} successfully learned expressive language models from web-scale data. The main missing characteristic of existing web-scale data for continual learning is the lack of naturally occurring \emph{tasks}\footnote{We use a definition of task similar to \citet{CLSurvey2}. A task is a set of data belonging to a specific domain, output space, or data distribution.}. In this paper, we propose to use social media, in which users post commentary in natural language. Specifically, we consider the popular social media platform Twitter. We define the problem of fitting a personalized language model to each user (\ie each user is a different task) and refer to this as Personalized Online Language Learning (\ourproblem). Users are added and dropped continuously; moreover, users post at different frequencies, which makes the task distribution highly non-stationary. This problem is continual, online, and multi-task. Learning the personalization of each user is an online multi-task learning problem, whereas learning the shared language model is a continual learning problem since the non-stationary user distribution changes the data distribution through time. In contrast to existing continual learning benchmarks, tasks are not disjoint and not sequenced as the data arrives in one continuous stream. This setting is sometimes called online continual learning.

Real data for \ourproblem is available on a massive scale and does not require extrinsic labeling as language learning is unsupervised. We collect a dataset of more than 100 million tweets with more than 1.5 billion tokens, posted by one million users over six calendar years\footnote{Publicly available at \url{https://github.com/firehose-dataset}}. We refer to this dataset as \ourdata.  Using \ourdata, we investigate architectures for \ourproblem and describe an effective baseline model that can be used to benchmark and develop continual learning algorithms.

\ourproblem, \ourdata, and our baseline architecture enable controlled evaluation of continual learning algorithms on web-scale data. We conduct such an evaluation and uncover deficiencies in current continual learning algorithms. In particular, the evaluated continual learning algorithms are significantly worse than their offline counterparts even when they are evaluated in the online setting (\ie on the most recent datapoints). Offline algorithms that fit all the data outperform continual learning algorithms even on the most recent datapoints, which continual learning methods are designed to fit best. This suggests that continual learning algorithms not only forget the past but also underfit the present, \ie a lack of plasticity is an issue in addition to a lack of stability. We conjecture that the fault is with the optimizer, and conduct a detailed investigation of gradient-based optimization in continual learning.

The standard optimizer for continual learning is online gradient descent with a fixed number of gradient steps at each time step. The number of gradient steps per iteration is critical as it directly affects learning and generalization. An excessive number of gradient steps hurts generalization, while an insufficient number of steps impairs learning. We propose to adaptively control the number of gradient steps using an online validation buffer. The key idea is to use part of the online data to measure generalization and adapt the optimization accordingly. We present a simple strategy to maintain this buffer without wasting data. The resulting continual gradient descent method (\ourmethod) outperforms prior continual learning schemes on \ourdata and other benchmarks.

In summary, our contributions include the following. (i)~A new problem setting for continual learning that features real data produced naturally on a massive scale, with natural temporal dynamics (\ourproblem). (ii)~A massive web-scale dataset (\ourdata) that can support research on \ourproblem. (iii)~An effective continual gradient descent algorithm (\ourmethod) that addresses the weaknesses of prior methods.

\section{Related Work}
% !TEX root = main.tex
Continual learning, also known as lifelong learning was first conceptualized by \citet{Thrun1}. It addresses the setting in which different but relevant tasks are learned sequentially, and poses the question \emph{can we enable inductive transfer between tasks?} This question has been studied within different contexts like lifelong robot learning, lifelong (semi-)supervised learning, lifelong reinforcement learning, and knowledge-based systems \citep{LifelongLearningBook}.

\citet{CatastrophicF} showed that neural networks manifest a behavior that is called catastrophic forgetting. This problem is studied in the general setting of the stability-plasticity dilemma \citep{stability_plasticity}, where models and algorithms are designed to control the tradeoff between stability (retaining knowledge) and plasticity (ability to learn). Following the recent popularization of deep learning, this problem has regained attention. The main thrust of continual learning research over the past few years specifically addresses the stability-plasticity dilemma in deep neural networks. Although the major focus of the lifelong learning literature has been the transfer of inductive bias from past to future tasks, 
while the focus of continual learning has been information retention,
we treat these two terms interchangeably since the same models, algorithms, and benchmarks can be utilized for both.

In this paper, we focus on the (self-)supervised lifelong/continual learning problem with deep neural networks and refer the interested reader to the book of \citet{LifelongLearningBook} for an extensive discussion on the literature of continual learning/lifelong learning. For the sake of consistency with the existing literature using deep neural networks, we use the name continual learning in the rest of the paper.

\paragraph{Continual learning methods.}
 Existing continual learning methods can be categorized into three major families based on how they store and use past data. \emph{Prior-based methods} leverage a data-driven prior from past tasks to regularize the current task~\citep{kirkpatrick2017overcoming,zenke2017continual,nguyen2017variational,ritter2018online,chaudhry2018riemannian,Schwarz2018:ICML,aljundi2019task, Zhang_Data, Jung_Data, Triki2017, Silver_Data, aljundi_data, Swaroop_2019, Ahn_2019, Chen_2019}. \emph{Replay-based methods} store a small set of samples as a proxy for the past, and leverage them together with the current task~\citep{shin2017continual,lopez2017gradient,chaudhry2018efficient,chaudhry2019continual, aljundi_er, isele_er, icarl, rolnick_er, robins_er, gdumb, buzzega2020dark}. \emph{Structure-based methods} isolate parameters of different task-specific components to prevent interference between tasks~\citep{rusu2016progressive,li2017learning,mallya2018packnet, Serra_Structure}. 
 
These methods are widely studied and empirically compared with each other using various modalities in recent continual learning surveys \citep{Parisi2019, CLSurvey2}. In this paper, we focus on the benchmarking of these methods from a data perspective. We specifically focus on \emph{online learning} aspects in which a learner is faced with a stream of data without any offline training or large batches of seperate tasks as recommended by \citet{CLSurvey2}. Moreover, we also aim to address the lack of complexity of the continual learning benchmarks as raised by \citet{Parisi2019}.  From a modelling perspective, we focus on replay-based methods as they have been shown to have superior performance on large-scale datasets while being computationally efficient~\citep{chaudhry2019continual,Parisi2019}.

\paragraph{Online learning.}
Online learning (OL) \citep{ShwartzBook, HazanBook} is closely related to continual learning. Online learning typically refers to the game-theoretic treatment of learning. In this game, the learner plays against an opponent which generates inputs and loss functions (labels) for the learner. In contrast to continual learning, the opponent is typically a strategic agent and learning is in an adversarial setting. Another common instantiation of online learning is playing against an opponent which is not strategic and generates identically and independently distributed (i.i.d.) datapoints, also known as online optimization. In contrast, continual learning is somewhere between these two settings and considers an opponent which is oblivious but time varying.  Most relevant to us is the non-convex OL setting of \citet{Hazan} as deep neural networks are highly non-linear. They propose a non-convex optimization method that can be seen as a special case of ours, as discussed in Section~\ref{subsec:oms}.

\paragraph{Multi-task NLP.} 
There has been remarkable progress in multi-task NLP due to recent breakthroughs in architectures and unsupervised pretraining~\citep{vaswani2017attention,gpt2,gpt3}, multi-task learning~\citep{Hashimoto2017,changpinyo2018multi,liu2019multi,houlsby2019parameter}, and comprehensive evaluation benchmarks~\citep{wang2018glue,Wang2019superglue}.
\citet{changpinyo2018multi} learned a set of task embeddings with a shared sequence model to address 11 sequence tagging tasks.
\citet{liu2019multi} trained a multi-task decoder on the Transformer model to jointly learn 10 different sentence-level NLP tasks.
\citet{houlsby2019parameter} further utilized residual adapters~\citep{RebuffiBV17} to adapt Transformers to different tasks.
Our work uses these multi-task architectures as a starting point and applies them to the massively multi-task continual setting of personalized language learning.

\paragraph{Controlled language generation.} Our problem setting can be seen as a form of controlled (conditional) language generation. Existing setups typically learn a generative model over a small set of control factors such as sentiments~\citep{hu2017toward,shen2017style}, sentence tense and length~\citep{ficler2017controlling,logeswaran2018content}, or content-derived topics and domains~\citep{lample2018multiple,dathathri2019plug}. A recent highlight is the work of~\citet{keskarCTRL2019}, which trains a very large Transformer with over 50 control codes, achieving high-quality text generation with attribute control. Compared to this body of work, we use a much larger number of control conditions (hundreds of thousands: each user is a control condition) in a continual streaming setting.

\paragraph{Web-scale continual learning.}
Lifelong learning systems has been studied extensively in the context of knowledge bases (KBs) for symbolic artificial intelligence. 
Although there were attempts to manually create these KBs~\citep{Cyc}, their massive scale deemed manual creation impractical. One of the first attemtps to learn these structured KBs is the NELL project by \citet{NELL}. NELL crawls the internet and continually learns natural language information; it is still functioning after a decade. Freebase \citep{freebase}, DBPedia \citep{dbpedia}, and Yago \citep{yago, yago2} implemented related ideas on a larger scale. While these KBs are language specific, NEIL \citep{NEIL} and Robobrain \citep{robobrain} covered multiple modalities. Following the popularity of deep learning approaches, these projects became less prominent. Our work is within the deep learning paradigm and does not use knowledge bases. We use unstructured Twitter data to learn neural language models in a continual setting. To the best of our knowledge, our work is the first attempt to benchmark recent continual learning ideas on web-scale data.

\section{Personalized Online Language Learning}
\label{sec:poll}
In this section, we formally define the personalized online language learning (\ourproblem) problem and discuss its properties.

Consider a stream of tweets generated by a large number of users over time. Our aim is to learn a personalized language model for each user. Language modeling is an unsupervised problem; however, we will discuss it in the supervised learning formalism since it is typically tackled as the problem of predicting the next word given a context. In other words, part of the sentence (tweet) is the input and the next word is the label. We denote the observed part of the tweet of the user $u$ at time $t$ as $\xx^{u}_t \in \mathcal{W}^{L}$ and the next word as $\yy^{u}_t \in \mathcal{W}$ where $\mathcal{W}$ is set of (sub)words and $L$ is the length of the context. We are interested in learning a parametric model $h(\cdot;\btheta^{sh},\btheta^u):\mathcal{W}^L \rightarrow \mathcal{W}$ which given context $\xx^u_t$, predict the next word $y^u_t$. Following the multi-task paradigm, we use models which share a subset of the parameters ($\btheta^{sh}$) between users and keep the rest user-specific ($\btheta^u$ for user $u$). In order to facilitate learning, we consider a loss function $l:\mathcal{W}\times \mathcal{W}\rightarrow\mathcal{R}^+$ which evaluates the accuracy of the predictions.

We model the Twitter stream as the following generative process. Users are distributed in a non-stationary way over time with distribution $\pi_t$ such that the set of $N_u$ users are sampled at time $t$ as $U_t \sim \pi_t^{\times N_u}$. Please note that some of these users have never been seen in the data stream, \ie users who just joined Twitter, while others have already been part of the network for some time. Each user then generates Tweets from their user-specific time-varying language model as $\xx_t^u,\yy_t^u \sim P_t^{u}$. We now summarize some of the properties of this setting.

\begin{itemize}
    \item \textbf{Multi-Task:} Each user's language model is a different task which shares significant commonalities, enabling the sharing of inductive biases.
    \item \textbf{Online:} Data is truly online and arrives as a stream. There are no task/user boundaries as users tweet spontaneously through time.
    \item \textbf{Continual/Lifelong:} New users are continuously added, requiring fast transfer. Moreover, some users have long periods of silence between data points, requiring learning without forgetting.  
\end{itemize}

To further constrain \ourproblem, consider the discrete streaming dataset over time $t$ as $\{\DD_t\}_{t \in 1\ldots,T}$ where each $\DD_t$ is a small batch of data $\DD_t = \{\xx_t^1,\yy_t^1,u_t^1\}, \ldots, \{\xx_t^{N_t},\yy_t^{N_t},u_t^{N_t}\}$ as set of tuples of context, next word, and user-id where $N_t$ is the batch size. For notational clarity, we define the empirical loss over a set of data points $\SSS$ as ${\lL^{\SSS}(\btheta) = \nicefrac{1}{|\SSS|\sum_{\xx,\yy,u \in \SSS} l(h(\xx;\btheta^{sh},\btheta^{u}),\yy)}}$. 
We only consider learners that can carry a limited amount of data (memory buffer $\MM_t$) across iterations.

There are two major objectives in \ourproblem: online learning and information retention. \emph{i)~Online learning} is evaluated as the loss on the next batch: $\lL^{\DD_t}(\btheta_{t-1})$. \emph{ii)~Information retention} is evaluated as the average loss on the previously seen data: $\frac{1}{t-1} \sum_{b=1}^{t-1}  \lL^{\DD_b}(\btheta_t)$. We are interested in minimizing these loss functions through time.

In summary, the learner plays the following $T$-step game with Nature at each time instant $t=1,\ldots,T$:

\begin{itemize}
    \item Nature determines $\pi_t$ and samples $\{\xx^i_t,\yy^i_t, u^i_t\}$ via $u^i_t \sim \pi_t$ and $\xx^i_t,\yy^i_t \sim \pP^{u^i_t}_t$
    \item Nature reveals $\{\xx^i_t, u^i_t\}$ \vspace{-1mm}
    \item Learner decides $\{\hat{\yy}^i_t = h(\xx^i_t;\btheta^{sh}_t, \btheta^{u^i_t}_t)\}$
    \item Nature reveals $\{\yy^i_t\}$ and learner pays both losses
    \item Learner sets $\btheta_{t+1}$ and $\MM_{t+1}$ using $\btheta_{t},\xx^i_t,\yy^i_t$, and $\MM_t$
\end{itemize}
To play this game successfully, the learner needs to remember the past using its model parameters and the limited memory buffer, fit the online stream as closely as possible, and generalize to the future.

\section{Firehose Datasets}
\label{sec:dataset}
% !TEX root = main.tex
In this section, we discuss the \ourdata datasets, which we collect to benchmark continual learning algorithms. We first discuss how we collected and processed web-scale data, and then analyze aspects of the dataset that demonstrate its applicability to continual learning.

\subsection{Data Collection and Preprocessing}

We collected a large dataset of Twitter posts, posted between January 2013 and September 2019. We perform the following steps of data cleaning. (1) Perform language detection and remove all non-English tweets. (2) Remove tweets that are explicitly marked as retweets and replies. (3) Remove users that have the total number of tweets less than 20. (4) Tokenize and replace all URLs as a special token ``\texttt{<URL>}''. (5) Prepend and append the meta-tokens ``\texttt{<SOT>}'' and ``\texttt{<EOT>}'' in front of and after each tweet. The preprocessing steps result in 110$M$ tweets in total from more than 920$K$ unique users. We then randomly split the collected data into two datasets, \ourdata10M and \ourdata100M, which contain 10$M$ and 100$M$ tweets, respectfully.
\ourdata10M is intended for rapid development and validation, while \ourdata100M is intended as a large-scale benchmark. We also prepare another non-overlapping set of data (with 100$M$ tweets) from Jan 2013 to Dec 2013 as the pretraining dataset, to learn a user-agnostic Transformer that models the Twitter language. We note that the purpose of this model is to serve as an initialization for continual learning of personalized Twitter language models. We set aside 3 randomly sampled tweets per user for the validation set, and 3 other randomly sampled tweets per user for the test set. We use the test set to compare continual learning with offline training, and to evaluate models in a time-invariant way.

The number of unique tokens in the data is far beyond the capacity of word-based models. To address this, we leverage practices from machine translation~\citep{wu2016google} and build a subword vocabulary through unigram modeling~\citep{kudo2018subword} using the SentencePiece library~\citep{kudo2018sentencepiece}. This produces a subword vocabulary of size 32,000. We use a subword vocabulary for training only, and report word perplexity for evaluation, since this measure is independent of different tokenizers and vocabularies.

\subsection{Properties of the \ourdata Datasets}

\begin{figure}[t]
	\centering
	\includegraphics[width=\textwidth]{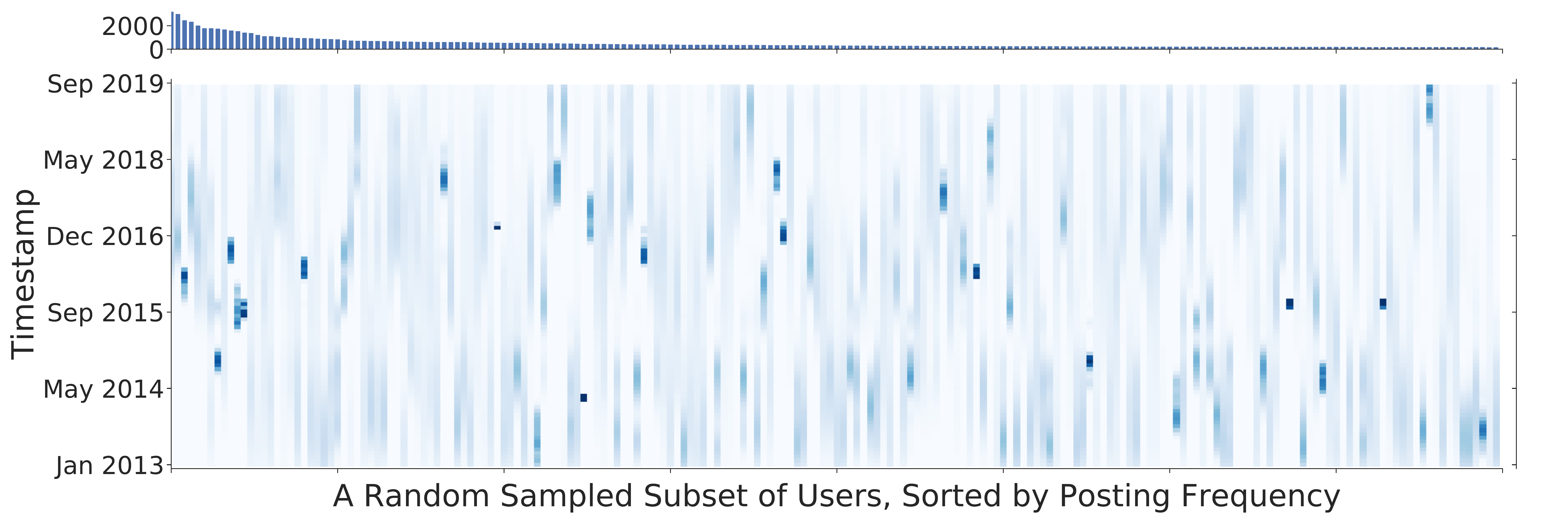}
	\vspace{-5mm}
	\caption{User activity distributions. The vertical axis shows the density of posts for each user. Darker means higher. The total number of posts per user is on the top. Distributions are heterogeneous and highly non-uniform.}
\label{fig:data_distribution}
\end{figure}

We summarize the key statistics of \ourdata in 
Table~\ref{tab:dataset}. Figure~\ref{fig:data_distribution} shows a small sample of the temporal dynamics in the data. The distributions of users' posts are continuous, span a long time horizon, and are highly heterogeneous.

\begin{table}[hbt]
    \centering
    \caption{Key statistics of the \ourdata datasets.}
    \begin{tabular}{@{}lrrr@{}}
        \toprule
         &  \# Users &  \# Tweets & \# Tokens \\ \midrule
        {Firehose10M}  & $94.0$K & $10.4$M & $173.3$M \\
        {Firehose100M} & $917.4$K & $100.4$M & $1672.7$M \\
        \bottomrule
    \end{tabular}
    \label{tab:dataset}
\end{table}

\begin{figure}[ht]
\centering
    \includegraphics[width=0.6\columnwidth]{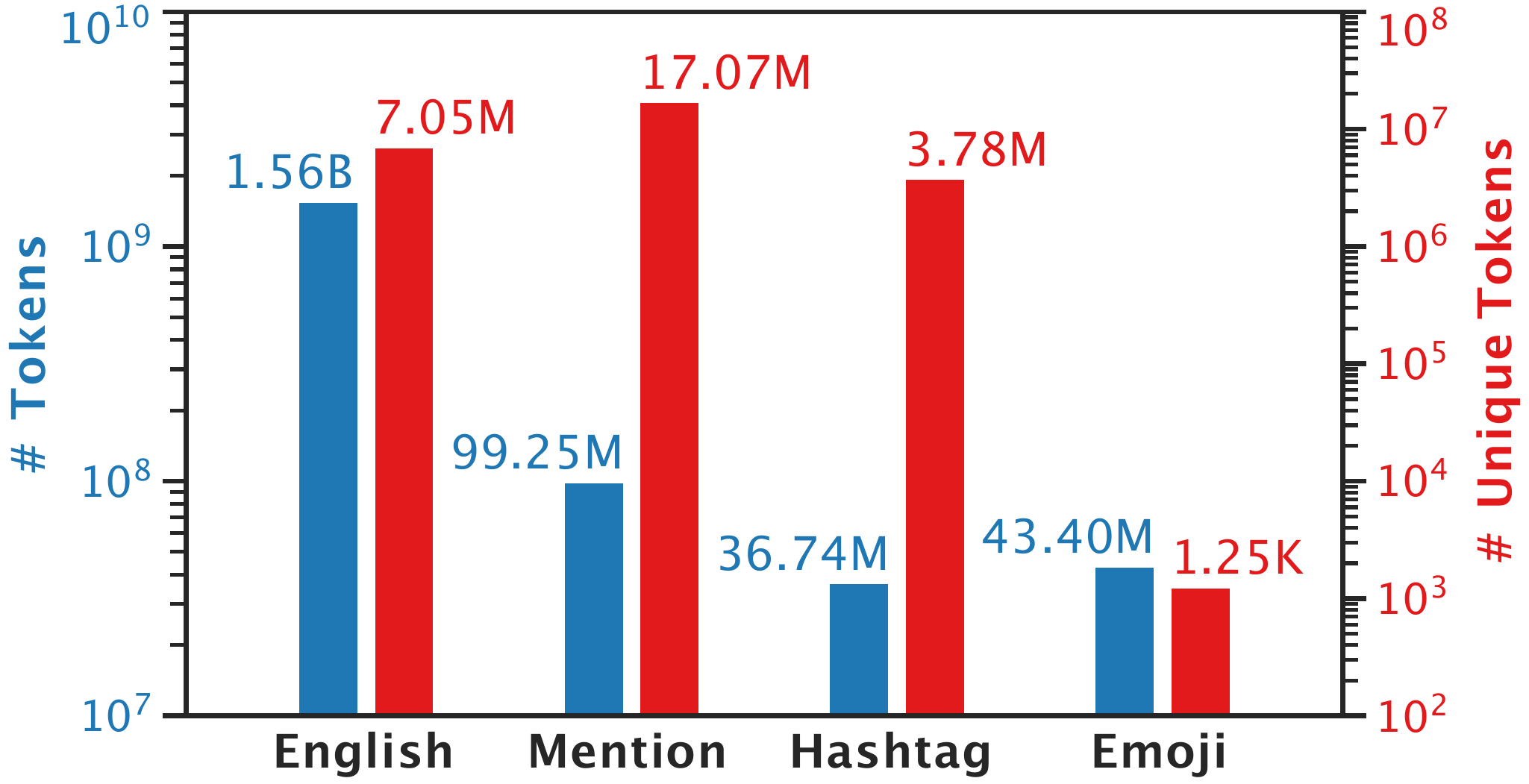}
    \caption{Frequency of tokens in \ourdata100M. The vertical axes are on logarithmic scales.}
    \label{fig:data_frequency}
\end{figure}

\begin{figure}[ht]
    \centering
    \includegraphics[width=\textwidth]{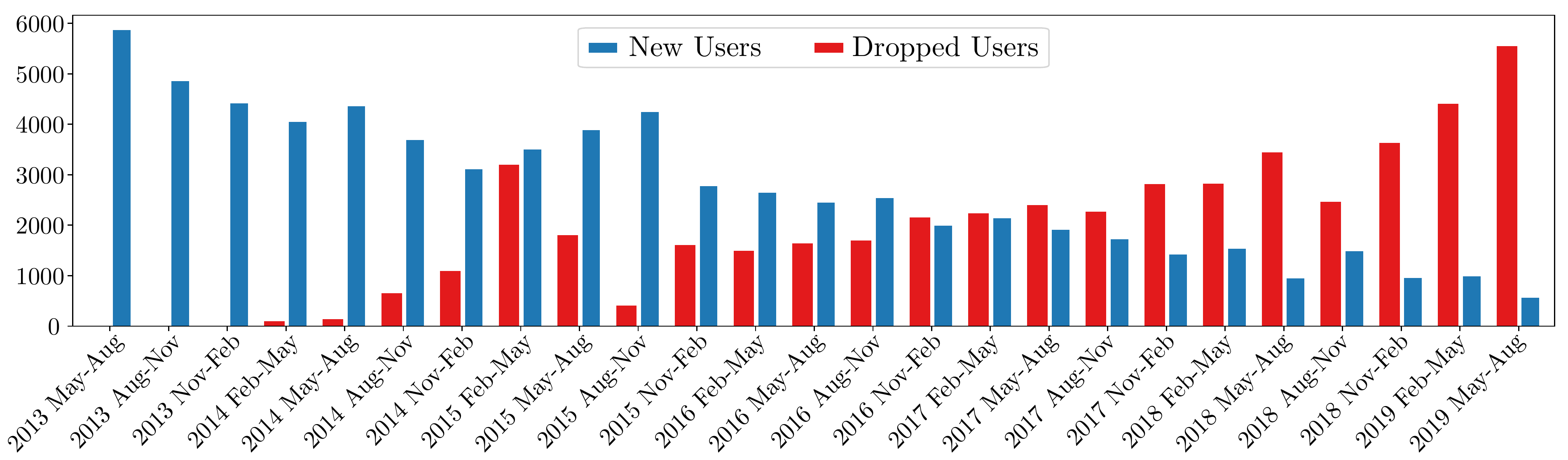}
    \vspace{-5mm}
    \caption{New users and users who become inactive over time. (Statistics reported on a random 10\% subset of {\ourdata100M}.)}
    \label{fig:add_drop}
\end{figure}

\begin{figure}[ht]
    \centering
    \includegraphics[width=0.9\textwidth]{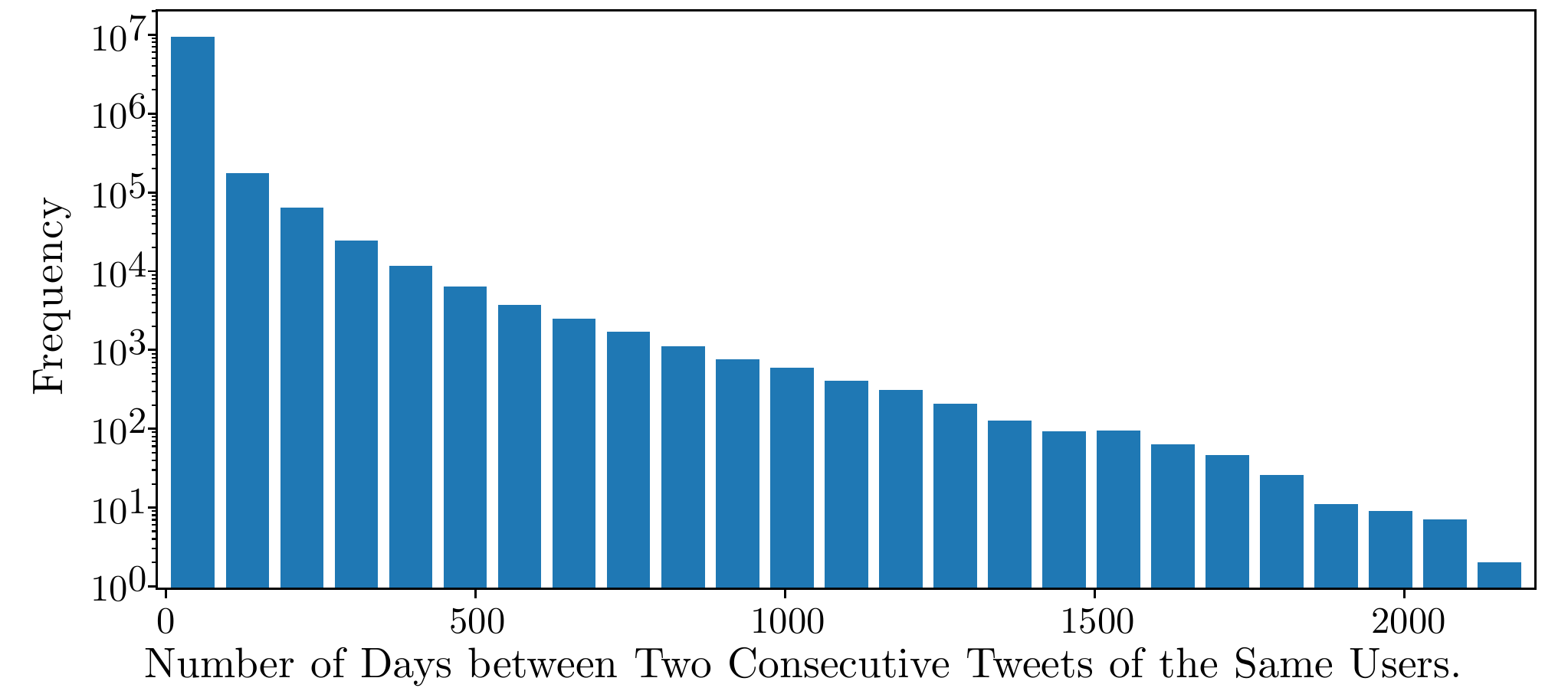}
    \caption{Histogram of the number of days between two consecutive tweets of the same user. There are many cases where users stay inactive for more than a year. Hence, information retention is critical in \ourdata. (Statistics reported on a random 10\% subset of {\ourdata100M}.)}
    \label{fig:hist_time_diff}
\end{figure}

Figure~\ref{fig:data_frequency} shows the distribution of word-level tokens in the \ourdata datasets, categorized into English, hashtags, mentions, and emoji. (A mention is a direct reference to another user.) In total, \ourdata10M and 100M contain 5.2M and 26.2M unique word-level tokens, respectively.
While English dominates in terms of the number of tokens, the number of unique tokens is dominated by mentions and hashtags. The high number of unique mentions and hashtags indicates that many of these only occur a small number of times. The emoji vocabulary is limited but heavily used: a higher number of tokens than hashtags, for a vocabulary that is three orders of magnitude smaller.

\ourproblem is online, multi-task, and continual. In order to check whether \ourdata is a good fit for \ourproblem, we need to validate these three aspects. The online and streaming nature of the data is clear from Figure~\ref{fig:data_distribution}. Next, we focus on the multi-task and continual aspects.

\begin{table}[h]
    \centering
    \caption{Word perplexity ($\downarrow$) of a personalized vs a user-agnostic model (\# parameters in parentheses) on {\textsc{Firehose}10M}.}
    \tabcolsep 3pt
    \begin{tabular}{@{}l@{}c@{\hspace{2mm}}c@{}}
    \toprule
	& {User-agnostic (57.4M)} & {Personalized (65.1M)} \\ \midrule
    Perplexity $\downarrow$ & $112.5$ & $\mathbf{94.1}$ \\ \bottomrule
    \end{tabular}
    \label{tab:mtl_models}
\end{table}

\paragraph{Multi-task structure.} We conduct an experiment to evaluate the importance of multi-task modeling on \ourdata. To this end, we examine whether a single language model can fit the data as well as a personalized language model. Table~\ref{tab:mtl_models} shows the performance of a user-agnostic language model versus a personalized (user-conditioned) model on {Firehose10M}. For this experiment, we trained the models offline\footnote{The models and training strategy are discussed in Section~\ref{sec:architecture}.}, in a batch setting, to specifically evaluate the prevalence of multi-task structure in the data. The personalized model reduces word-level perplexity by 16.3\% relative to the user-agnostic model. To further evaluate the multi-task structure, we evaluate each user's language model on data from $5$ other randomly chosen users. Cross-user perplexity is as high as $159.2$. This indicates that tasks (users in our setting) differ significantly from each other, validating the multi-task structure. 

\paragraph{Continual structure.} Continual learning has two key aspects: forward transfer and information retention. In order to analyze the presence of these two aspects in \ourdata, we plot the number of new users and dropped users for each three-month period in Figure~\ref{fig:add_drop}. New users are those that just joined Twitter; dropped users are ones that stop tweeting at this time. As shown in Figure~\ref{fig:add_drop}, there are thousands of new users which need quick adaptation/forward transfer. (The decreasing and increasing trends are due to our filtering of users before dataset creation.) In order to show the need for information retention, we plot the histogram of the duration between each consecutive tweet of users in Figure~\ref{fig:hist_time_diff}. For many users, there are long periods of silence, some as long as years. Thus models must be capable of retaining information for such long periods of time.

\section{Model Architectures for \ourproblem}
\label{sec:architecture}
%Most continual learning algorithms are applicable regardless of the model and data. Hence, it is important to provide the right dataset and model architecture for fair and efficient evaluation as a complete benchmark. We just proposed \ourdata as the data part of the benchmark.

In this section, we study existing modeling choices and provide recommendations for effective and efficient models for \ourproblem. We perform a study over multiple architectural choices. We first introduce the principles and concrete instantiations of architectures for \ourproblem, and then discuss the experimental results and identify the best architectures.

\subsection{Multi-task Learning Architectures for \ourproblem}
\label{subsec:supp:mtl_model}

An architecture for \ourproblem requires a high-capacity sequence model that excels at learning linguistic patterns within a long context in order to model the language, and a highly scalable personalization mechanism that can adapt the model to a large number of individuals. As a high-capacity shared model, we use Transformer-XL~\citep{dai2019transformer}. Based on it, we test a variety of approaches from the multi-task learning literature to encode user-specific context that the Transformer-XL can be conditioned on. For scalability to a high number of users, we only consider mechanisms that capture user-specific information as low-dimensional embeddings. This also allows us to extend to new users, as adding a new user only requires a new embedding.

\begin{figure}[t]
    \centering
    \includegraphics[width=0.975\columnwidth]{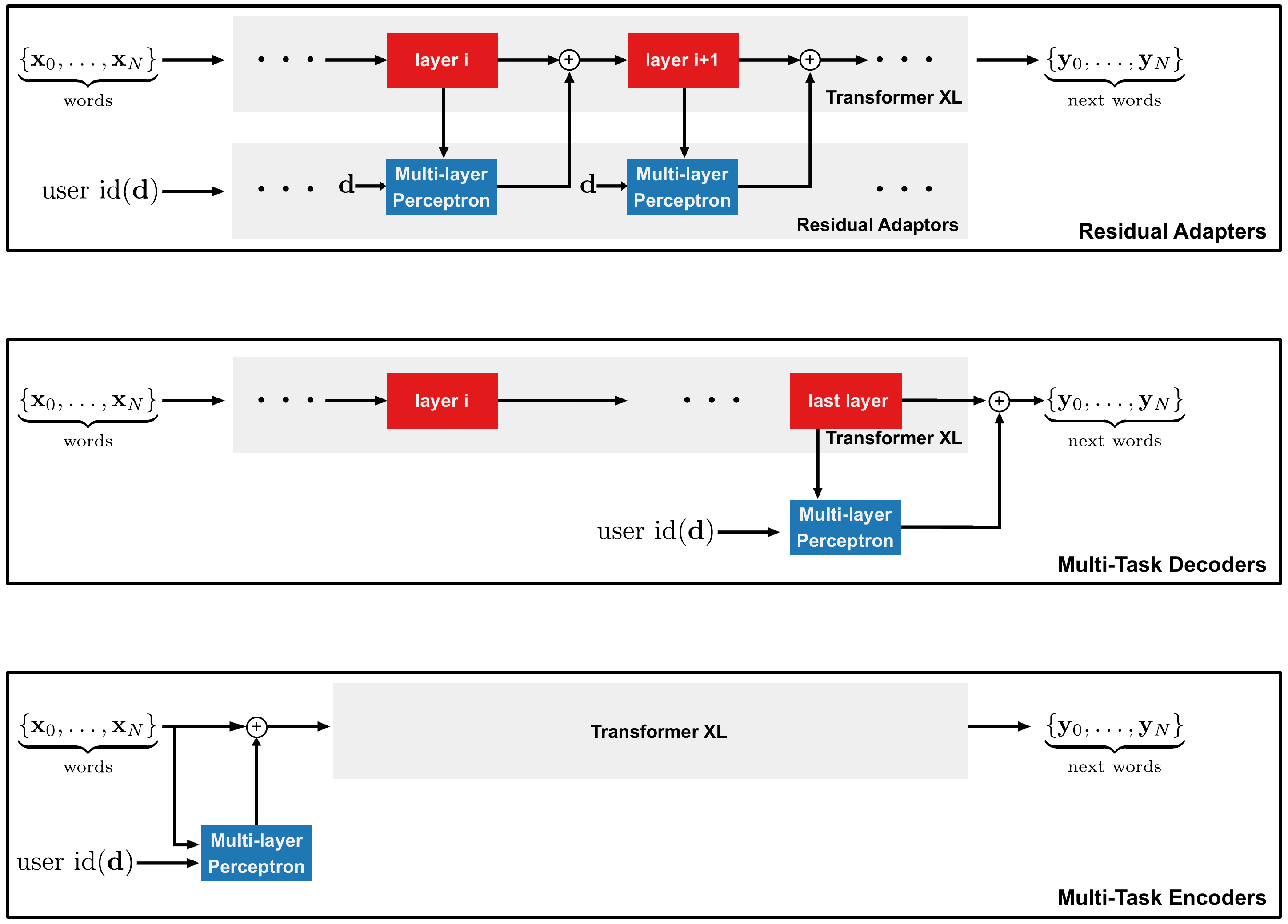}
    \caption{Illustration of multi-task transformer architectures for \ourproblem.}
    \label{fig:supp:mtl_model}
\end{figure}

We consider three different categories of multi-task architectures (see Figure~\ref{fig:supp:mtl_model} for illustrative examples). (1) \emph{Multi-task Encoder} is a model that combines the user-specific information together with the word embedding before any sequence modeling happens. It uses a residual MLP to extract such user-specific word embedding, which is then provided to the shared sequence model. (2) \emph{Multi-user Decoder} uses the user embedding and a residual MLP to modulate the contextualized output embedding of the sequence model, akin to multi-head models. (3) \emph{Residual Adapters} is a layer-wise deep multi-task model \citep{RebuffiBV17,houlsby2019parameter}. It makes use of the user embeddings to modulate the output activation of each layer in a deep sequence model (Transformer-XL in our case) by predicting the user-specific residual component of intermediate representations.

\subsection{Quantitative Comparison of Different Architectural Choices}
\label{subsec:supp:mtl_result}
We instantiate the aforementioned options by using the 12-layer base Transformer-XL (with the exact sample configuration of \citet{dai2019transformer}). There is one residual MLP for both \emph{Multi-task Encoder} and \emph{Multi-task Decoder}, which takes the concatenation of the user embedding and the word embedding (or Transformer-XL's output activation) as input and outputs the user-specific residual component. The dimensionality of the hidden layers for this MLP is $128$. The \emph{Residual Adapters} model has 12 such residual MLPs (one corresponds to each Transformer-XL layer), with their hidden dimensionality also $128$. We list the capacity of each architecture in Table~\ref{tab:supp:mtl_models}.

We compare these architectures after an online-to-offline conversion as our focus here is on the model architectures, not continual learning. These models are learned offline with multiple epochs over the \ourdata10M data and evaluated on the balanced test set from all users. Table~\ref{tab:supp:mtl_models} reports the average perplexity over the balanced test set.

\begin{table}[htb]
    \centering
    \caption{Offline perplexity of multi-user architectures on \ourdata10M.}
    \vspace{1mm} 
    \tabcolsep 3pt
    % \resizebox{\columnwidth}{!}
    {%
    \begin{tabular}{l@{\hspace{1cm}}r@{\hspace{1cm}}r}
        \toprule
        Architectures & \# Parameters & Perplexity $\downarrow$ \\ \midrule
        {User-agnostic Language Model} & $57.4${M} & $112.5$ \\ \midrule
        \multicolumn{2}{@{}l}{Personalized Language Models} \\
         \hspace{2mm} Multi-task Encoder & $60.8$M & $100.6$ \\
         \hspace{2mm} Multi-task Decoder & $60.8$M & $101.4$ \\
         \hspace{2mm} Residual Adapters & $65.1$M & $\mathbf{94.1}$ \\
        \bottomrule
    \end{tabular}}
    \label{tab:supp:mtl_models}
\end{table}

The results suggest that personalization is possible with all of the aforementioned architectures. Furthermore, \emph{Residual Adapters} significantly improve perplexity over other architectural choices for a modest increase in the number of parameters. Hence, in our setting, the \emph{Residual Adapters} architecture is an effective choice. We use it as the standard model for subsequent experiments.

\section{Continual Gradient Descent}
\label{subsec:oms}
In gradient descent (GD), the number of iterations per batch is directly relevant to generalization. A higher number of GD steps increases algorithmic complexity and decreases generalization, as discussed by \citet{Schmidhuber97} from a Kolmogorov complexity perspective and by \citet{Yao2007early} for reproducing kernel Hilbert spaces. We analyze the role of the CL optimizer in this light.

At time $t$, given the observed data $\DD_{t}$ and the replay buffer $\MM_t$, the learner must determine the parameters $\btheta_{t+1}$. The information retention loss $\big(\frac{1}{t} \sum_{s=1}^{t} \lL^{\DD_s}(\btheta_{t+1}) \big)$ and the online learning loss $\big(\Ee_{\DD_{t+1}} [\lL^{\DD_{t+1}}(\btheta_{t+1})]\big)$ can be decomposed as follows:
\begin{align}
    \frac{1}{t} \sum_{s=1}^{t} \lL^{\DD_s}(\btheta_{t+1})&=  \underbrace{\frac{1}{t} \sum_{s=1}^{t} \lL^{\DD_s}(\btheta_{t+1})-\lL^{\overline{\MM}_t}(\btheta_{t+1})}_{\text{core-set loss}} 
    +\underbrace{\lL^{\overline{\MM}_t}(\btheta_{t+1})}_{\text{training loss}} \label{eq:dec1} \\
    \Ee[\lL^{\DD_{t+1}}(\btheta_{t+1})] &= \underbrace{ \Ee[\lL^{\DD_{t+1}}(\btheta_{t+1})] - \lL^{\overline{\MM}_t}_{t}(\btheta_{t+1})}_{\text{generalization}}  +  \underbrace{\lL^{\overline{\MM}_t}(\btheta_{t+1})}_{\text{training loss}} \label{eq:dec2}
\end{align}
where \mbox{$\overline{\MM}_t = \MM_t \cup \DD_t$} denotes combination of the replay buffer and the data from time $t$. 

We refer to the difference between the loss over the replay buffer and the history as the core-set loss in~\eqref{eq:dec1} since the replay buffer functions as a `core set' that summarizes past data. As long as the replay buffer is chosen effectively, the core-set loss will be small for any $\btheta_{t+1}$, making forgetting behavior independent of the chosen $\btheta_{t+1}$. Thus the optimizer can drive the training loss to convergence without concern for catastrophic forgetting.

Optimization for the online fit in~\eqref{eq:dec2} is trickier because it requires careful control of complexity. Consider taking multiple GD steps over the loss in~\eqref{eq:dec2} in order to find $\btheta_{t+1}$. With a higher number of iterations, the training loss will decrease. On the other hand, each iteration is expected to increase algorithmic complexity and generalization error. Consider two extremes. 1) Performing a single GD step (common practice for CL): \mbox{$\btheta_{t+1}=\btheta_{t} - \eta \nabla \lL^{\overline{\MM}_t}(\btheta_t)$}. 2) Optimizing the current fit to the stationary point by repeated GD steps until $\|\nabla \lL^{\overline{\MM}_t}(\btheta_{t+1})\|_2 \leq \delta$ for a tolerance parameter $\delta$. (This corresponds to smoothed online GD, proposed and analyzed by \citet{Hazan}.) The theory \citep{Schmidhuber97,Yao2007early} indicates that the first option will result in a high training loss and good generalization, while the second option will result in a low training loss and poor generalization.

We propose an adaptive strategy to find the right trade-off between the two options dynamically in a streaming setting. The key idea is to use the validation loss in an online manner. We keep an online first-in first-out (FIFO) validation buffer ($\VV_t$) and modify the game as follows. Given the data ($\{\xx^i_t, u^i_t\}$) at time $t$, the learner makes predictions and pays the online fit loss. Then, the learner pushes the new data to $\VV_t$ and pops 
$\{\overline{\xx}^i,\overline{\yy}^i, \overline{u}^i\}$ from the validation buffer. Finally, the learner sets the next model and replay buffer ($\btheta_{t+1}$ and $\MM_{t+1}$) using $\btheta_{t},\{\overline{\xx}^i,\overline{\yy}^i, \overline{u}^i\}$, and $\MM_t$.

\begin{figure}[t]
\centering
\includegraphics[width=0.75\columnwidth]{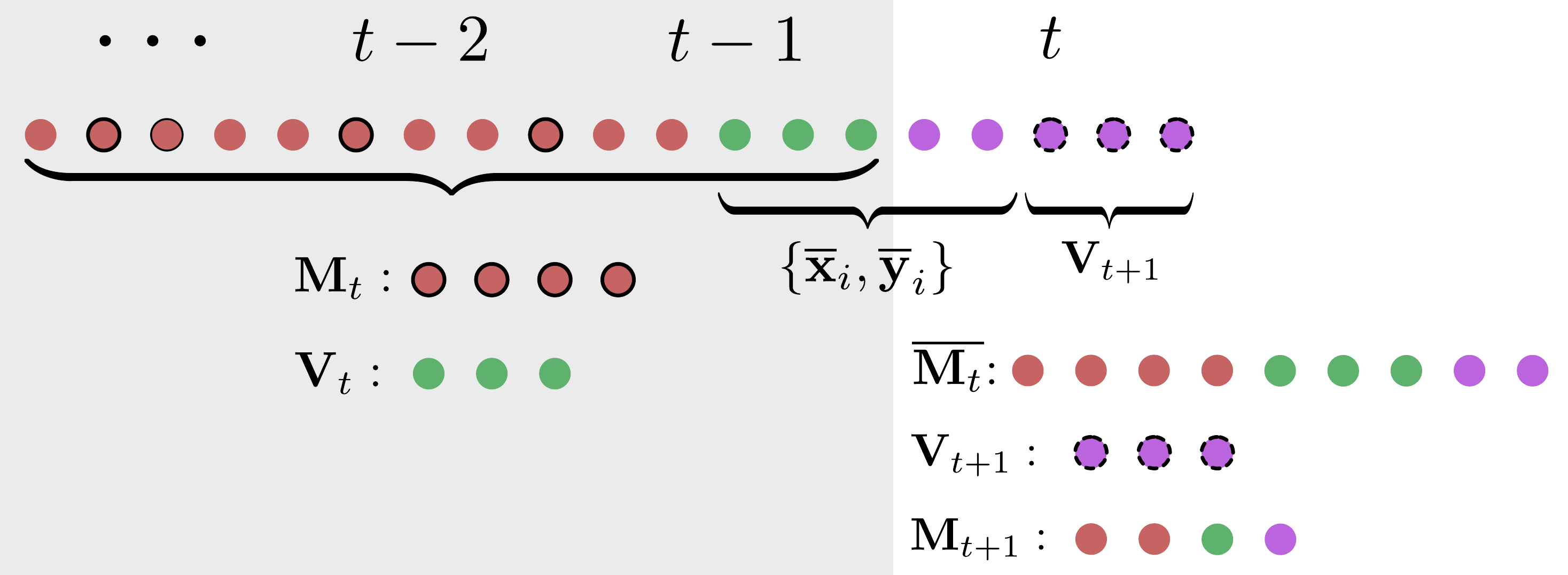}
\vspace{-5mm}
\caption{Illustration of the validation buffer with size 3. Circles are data points. \emph{Red:} data points used for gradient computation in the past. \emph{Green:} data points in the validation buffer. \emph{Purple:} New data that arrives at time $t$. Replay buffer is the circles with solid border as subset of the past data. Latest data points become new validation set, and all points except validation ones are used for gradient computation.}
\label{fig:data_flow}
\end{figure}

In this setting, the online-fit loss is computed using the current data. However, GD does not use it to compute gradients. Instead, new data is first pushed to the validation buffer. Gradient descent is performed on a loss function computed over the replay buffer and the data points from the previous iteration popped from the validation buffer. We visualize this streaming behavior in Figure~\ref{fig:data_flow}.

In order to choose the number of steps, we first perform the highest number of iterations we can afford and choose the one with the lowest validation error. Formally, we start with $\btheta_{t+1}^0 = \btheta_t$ and perform a sequence of gradient updates using the data \mbox{$\overline{\MM}_t=\{\overline{\xx}_i,\overline{\yy}_i\} \cup \MM_t$} as
\begin{equation}
    \btheta_{t+1}^k = \btheta_{t+1}^{k-1} - \eta \nabla \lL^{\overline{\MM}_{t}}(\btheta_{t+1}^{k-1}) \text{ for } k = 1,\ldots,K.
\end{equation}
The next iterate is chosen as $\btheta_{t+1}=\argmax_{\btheta \in \btheta_{t+1}^{0\ldots,K}} \lL^{\VV_{t}}(\btheta)$. Algorithm~\ref{alg:oms} summarizes our Continual Gradient Descent (\ourmethod) algorithm.

\begin{algorithm}[H]
\caption{\ourmethod: Continual Gradient Descent}
\label{alg:oms}
\begin{algorithmic}[1]
\State Initialize $\btheta_1$, $\MM_1$, $\texttt{ValidationBuffer}$.
\For{$t=1, \ldots, T$}
\State Observe $\{\xx_i\}$ and predict  $\{\hat{\yy}_i\}=\{h(\xx_i;\btheta_t)\}$ \Comment{Obtain new batch and predict next word}
\State Receive label $\yy_i$. \Comment{Compute and pay the losses}
\State $\texttt{ValidationBuffer.push}(\{\xx_i, \yy_i\})$  
\State $\{\overline{\xx}_i,\overline{\yy}_i\} = \texttt{ValidationBuffer.pop()}$
\State Set $\overline{\MM}_t = \{\overline{\xx}_i, \overline{\yy}_i\} \cup \MM_t$
\For {$k=1,\cdots,K$} \Comment{Compute candidate models for each $k$ GD steps}
\State $\btheta_{t+1}^{k}=\btheta_{t+1}^{k-1} - \eta \nabla {\lL}^{\overline{\MM}_t} (\btheta_{t+1}^{k-1})$
\EndFor
\State $\VV_t = \texttt{ValidationBuffer.peek()}$ \Comment{$\VV_t$ is not popped and stays in the buffer}
\State $\btheta_{t+1} = \argmin_{\btheta_{t+1}^{1},\ldots, \btheta_{t+1}^{K}} \lL^{\VV_{t}}(\btheta)$ \Comment{Choose the model with minimum validation error}
\State $\MM_{t+1} = \texttt{ReplayBufferUpdate}(\MM_{t},\DD_{t})$
\EndFor
\end{algorithmic}
\end{algorithm}

\section{Experimental Results}
Two aspects of continual learning approaches are the algorithmic model and the optimizer. For the algorithmic model, we experiment with the following. \emph{(i) Offline (Oracle):} Learner after online-to-offline conversion. This is not a continual learner as it passes over the data multiple times. We include it as an oracle model to compare against. \emph{(ii) Online Only:} Pure online learner without any memory. \emph{(iii) Replay Only:} An experience-replay-based method that adds all the data to a replay buffer and samples from it. \emph{(iv) Mixed Replay}~\citep{chaudhry2019continual}: Combines a pure online learner and a replay-based learner to take half of the data from the online stream and the other half from the replay buffer. \emph{(v) A-GEM}~\citep{chaudhry2018efficient}: An efficient implementation of Gradient Episodic Memory \citep{lopez2017gradient}. For the optimizer, we benchmark {Online GD} (i.e., applying SGD in an online manner) and \ourmethod. We evaluate each algorithmic model $\times$ optimizer pair.
%We report three metrics: offline test performance, online fit, and backward transfer.

In all experiments, we use the 12-layer Transformer-XL base model~\citet{dai2019transformer} as the backbone architecture for sequence modeling. We pretrain this model on a non-overlapping set of 2013 Twitter data (without using any user information) to serve as initialization for all the experiments (including the study of different architecture choices in Section~\ref{subsec:supp:mtl_result}). We set the dimension of user embeddings to 32, which amounts to 2.8M and 28.0M parameters for \ourdata10M and \ourdata100M, respectively. The subword vocabulary is instantiated with the dimension of 512, which aligns with the settings of the original Transformer-XL.

For replay-based continual learning algorithms, we use a small replay buffer (per user) to store historical data for all the replay-based methods we studied, which is 5 data points per user for \ourdata10M and 2 data points per user for \ourdata100M. For all the results that use the proposed online validation buffer, we allocate one additional data point per user, which incurs a limited memory cost.
We use the Adam optimizer~\citep{kingma2014adam} with a constant learning rate of $2.5\mathrm{e}{-4}\rm$. A warm-up strategy is applied, which linearly increases the learning rate from 0 to $2.5\mathrm{e}{-4}\rm$ in the first 2000 iterations. We also clip the gradient when its $\ell_2$-norm is greater than 0.25 to stabilize the training procedure in all experiments.
To facilitate reproducibility, we make our code publicly available at \url{https://github.com/firehose-dataset/congrad}.

\subsection{Online Learning}

In order to test online learning performance, we evaluate the next batch loss (online loss) as well as the performance of the final model.

\begin{figure}[htb]
    \centering
    \tabcolsep 1pt
    \begin{tabular}{cccc}
       \multicolumn{4}{c}{\includegraphics[width=0.6\columnwidth]{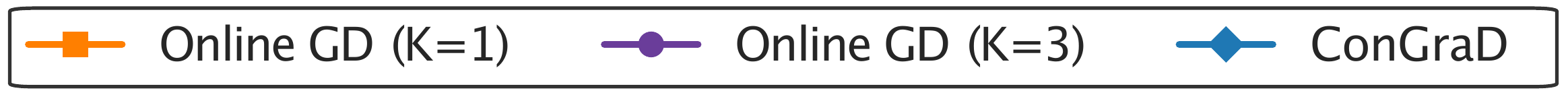}}\\
        \includegraphics[width=0.245\columnwidth]{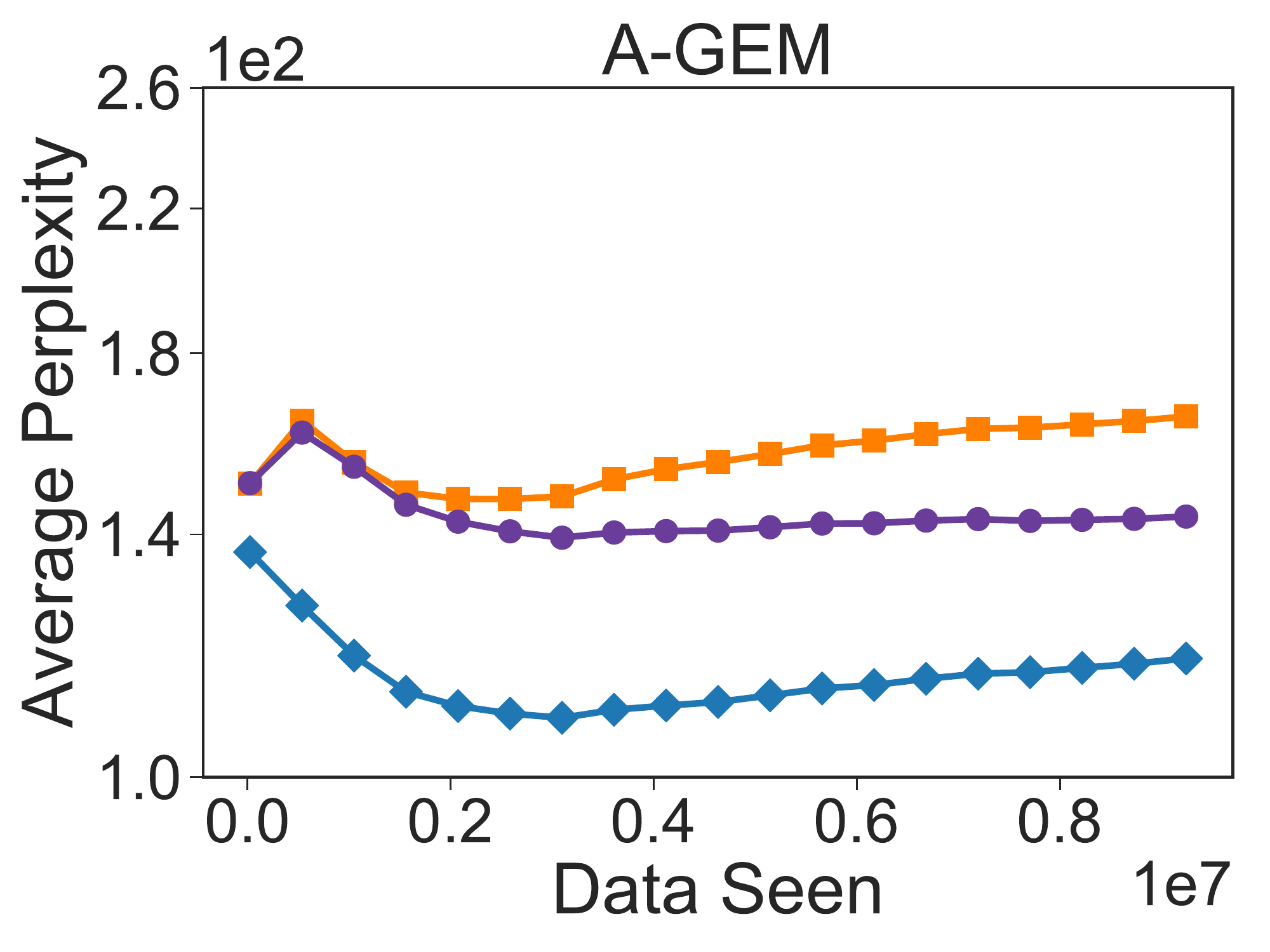} &
        \includegraphics[width=0.245\columnwidth]{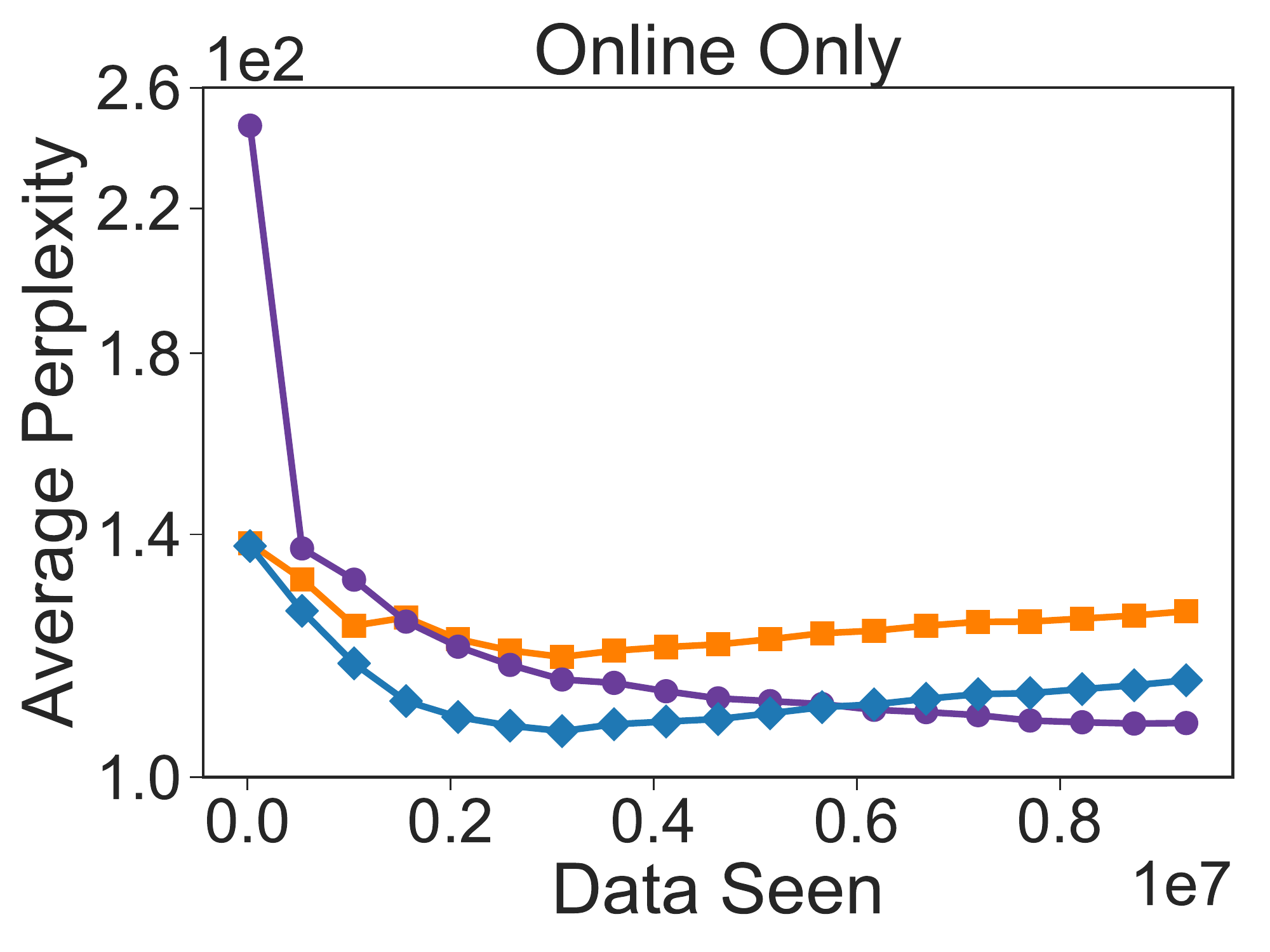} &
        \includegraphics[width=0.245\columnwidth]{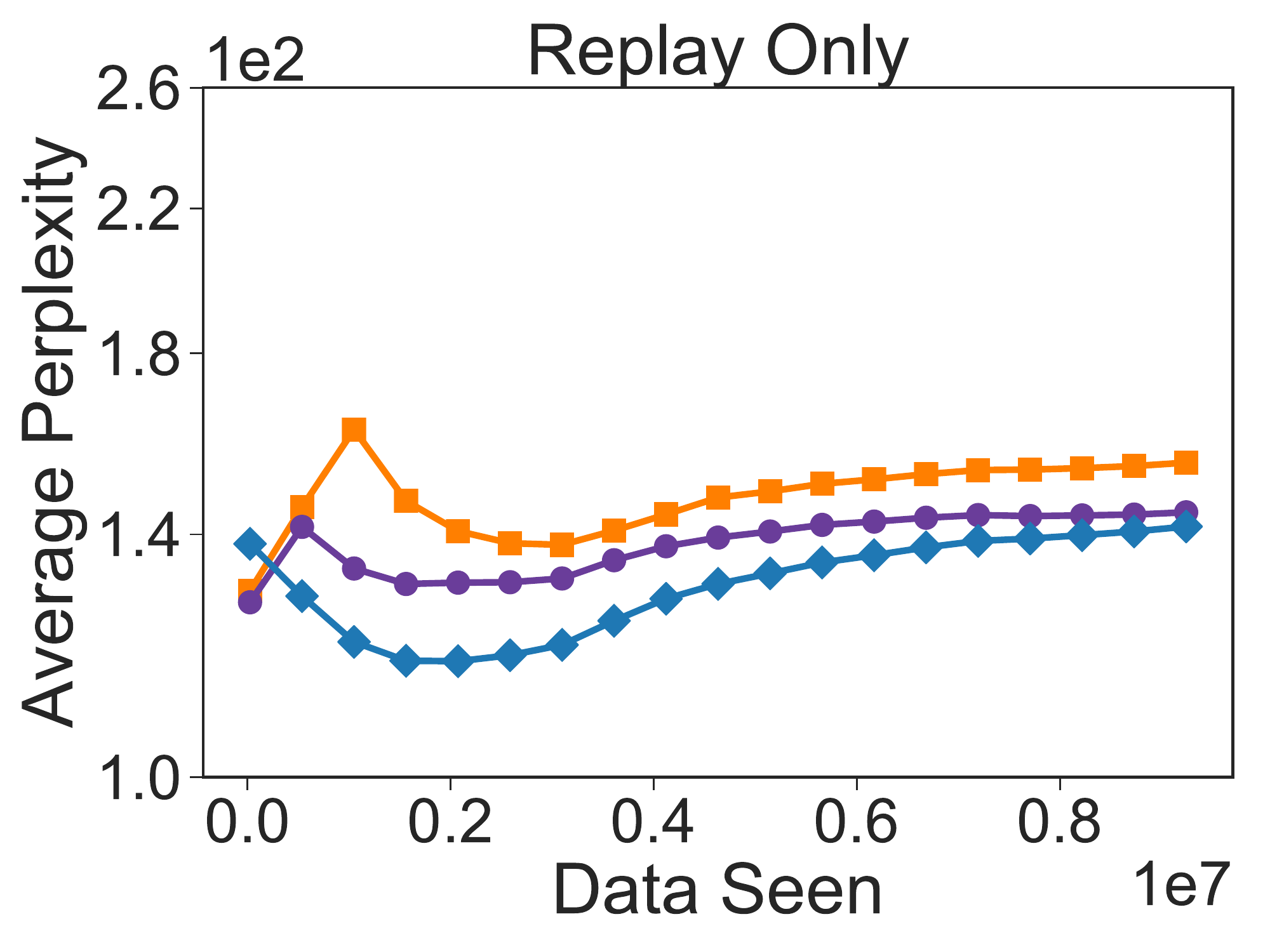} &
        \includegraphics[width=0.245\columnwidth]{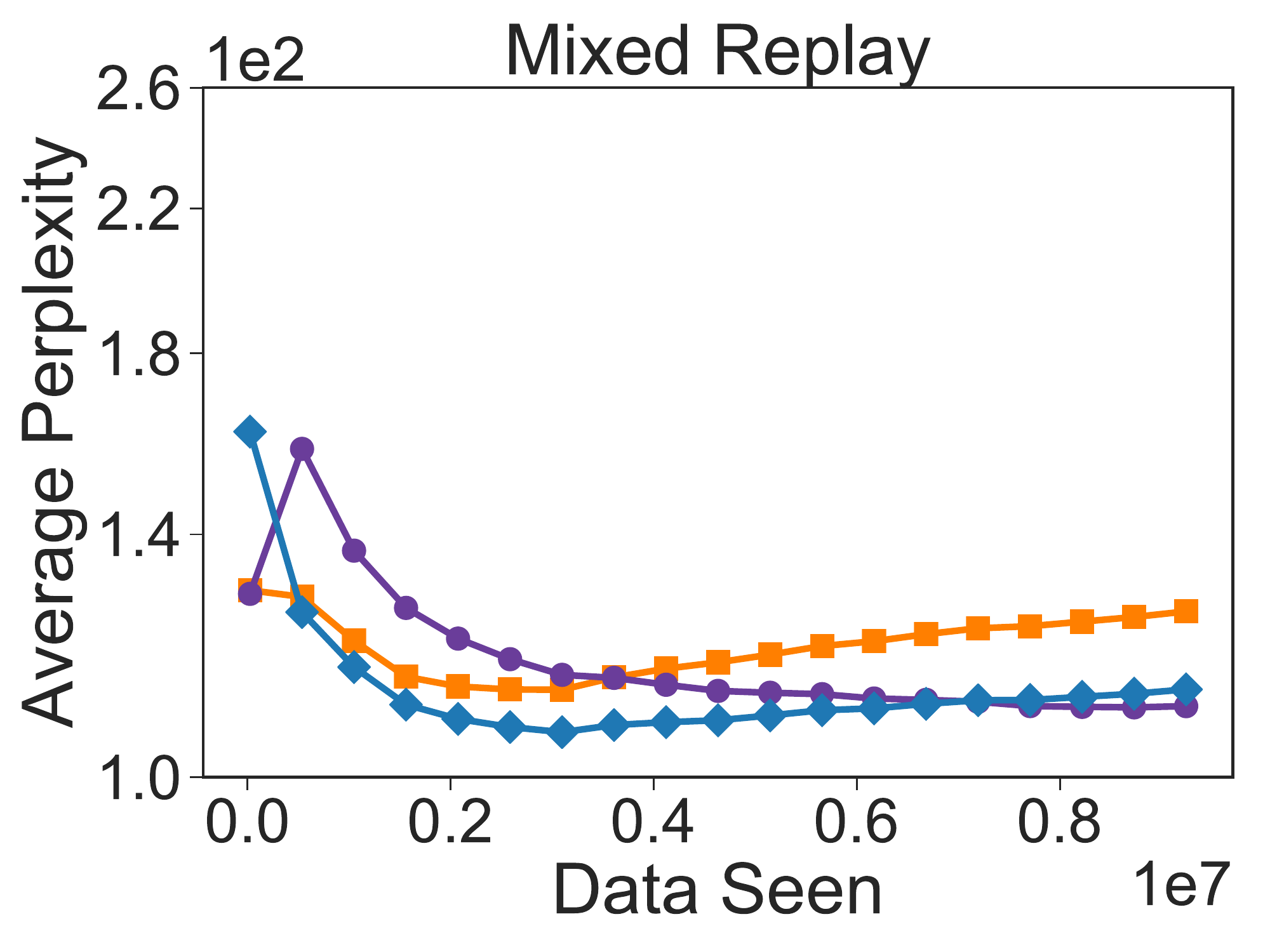}\end{tabular}
    \vspace{-2mm}
    \caption{\textbf{Online fit} though time. We plot average word perplexity of the next batch ($\downarrow$) through time.}
    \label{fig:online}
\end{figure}

For the online loss, we report word perplexity of the next (unseen) batch during training, as defined in Section~\ref{sec:poll}. We average it over time as $\frac{1}{t-1}\sum_{s=1}^{t-1} \lL^{\DD_{s+1}}(\btheta_{s})$ and plot over $t$ in Figure~\ref{fig:online}. This is similar to average regret in online learning. {Online GD} in an online-only setting is provably optimal for this metric, as shown by \citet{Hazan}. \ourmethod performs nearly as well as {Online GD} in the online-only setting and outperforms it in other settings. In terms of algorithmic models, A-GEM and Replay Only underperform as they focus exclusively on history without consideration for online fit. This suggests that if information retention is not considered, pure online learning works best, as expected. In terms of optimizers, both \ourmethod and {Online GD} perform well. \ourmethod learns faster but has some regret due to the finite size of the validation buffer. {Online GD} has no regret but learns slower. If other algorithmic settings are utilized for better information retention, \ourmethod is the best optimizer.

In order to evaluate the generalization ability of the final learned model, we report word perplexity on a withheld test set that is balanced across users. This is a time-independent measure over the full dataset which is comparable to offline training. It measures generalization and the online-offline gap. Table~\ref{tab:test_result} reports the test performance of each algorithm for \ourdata10M and \ourdata100M.

\begin{table}[htb]
\centering
\caption{ \textbf{Test performance.} We report word perplexity ($\downarrow$) of the final model on an unseen test dataset which is balanced across users.}
\label{tab:test_result}
\begin{tabular}{l@{\hspace{4mm}}c@{\hspace{2mm}}c@{\hspace{4mm}}c@{\hspace{2mm}}c}
\toprule
    & \multicolumn{2}{c}{\ourdata10M} & \multicolumn{2}{c}{\ourdata100M} \\
     \cmidrule(lr){2-3} \cmidrule(l){4-5} 
    & { {Online GD} } & { \ourmethod }  & { {Online GD} } & { \ourmethod } \\
    \midrule
    A-GEM &   $149.7$ & $\mathbf{148.1}$ & N/A & N/A \\
    Online only &  $153.1$ & $\mathbf{145.7}$ & $133.0$ & $\mathbf{125.9}$ \\
    Replay only & $142.0$ & $\mathbf{139.3}$ & $132.0$ & $\mathbf{128.2}$ \\
    Mixed replay & $133.3$ & $\mathbf{126.5}$ &  $120.2$ & $\mathbf{115.6}$ \\ \bottomrule
\end{tabular}
\end{table}

\ourmethod outperforms {Online GD} for all algorithmic models. This indicates that \ourmethod is a better optimizer from a test performance perspective. A-GEM does not scale to \ourdata100M; hence, it is only included for \ourdata10M. Mixed replay performs best among the modeling choices, in agreement with \citet{chaudhry2019continual}. The oracle offline learner attains $94.1$ perplexity, which is significantly better than the best-performing continual learner (Mixed Replay with \ourmethod). Although part of this gap will remain because online learning is provably harder than offline, we conjecture that significant progress is possible. Moreover, the role of scale is very significant: an 11-point difference in perplexity between \ourdata10M and 100M, validating the importance of evaluating continual learning on a large scale.

\subsection{Information Retention}

\begin{figure}[t]
%    \vspace{-5mm}
    \centering
    \tabcolsep 1pt
    \begin{tabular}{cccc}
       \multicolumn{4}{c}{\includegraphics[width=0.6\columnwidth]{files/legend.pdf}}\\
         \includegraphics[width=0.245\columnwidth]{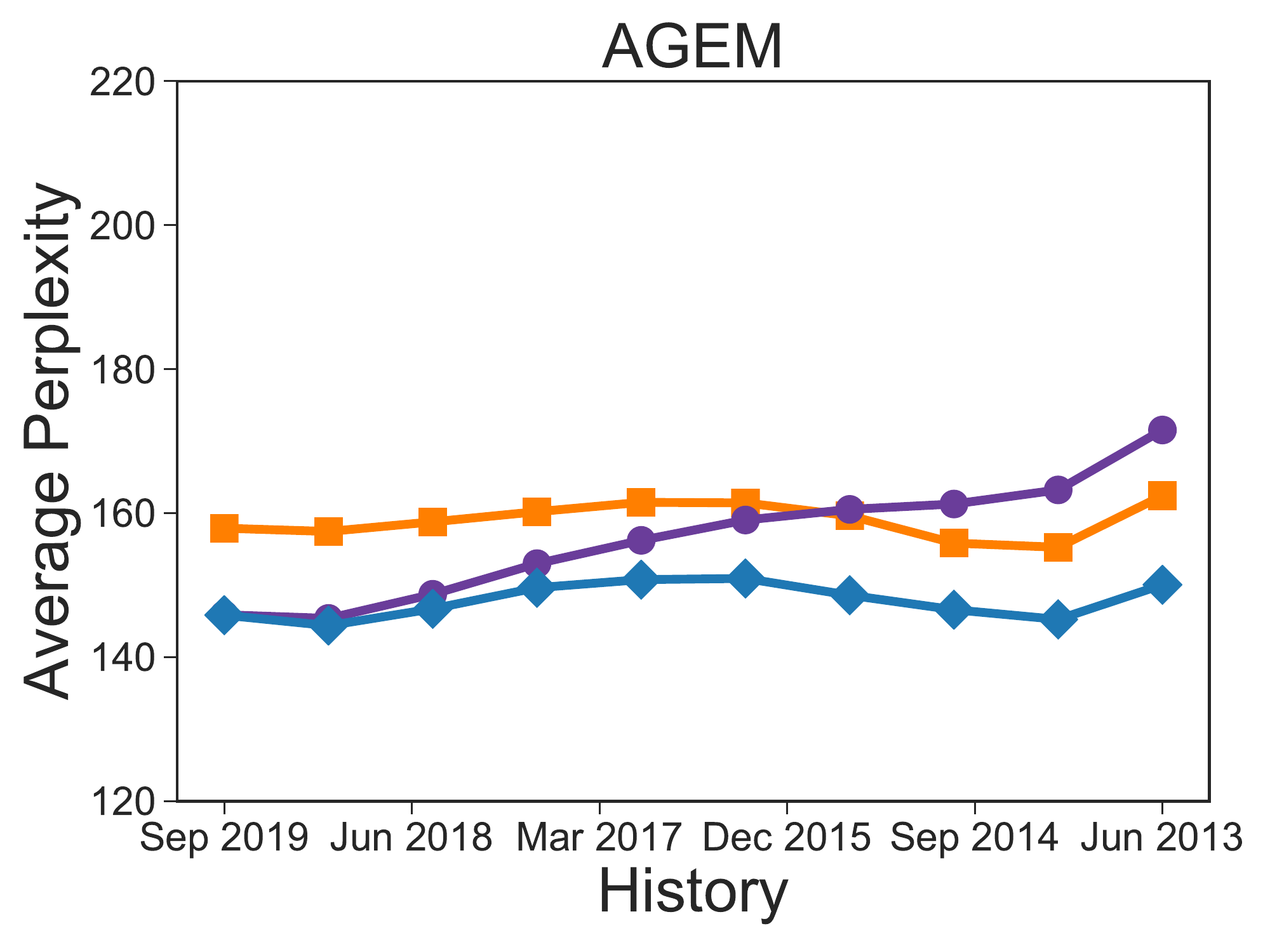} & 
        \includegraphics[width=0.245\columnwidth]{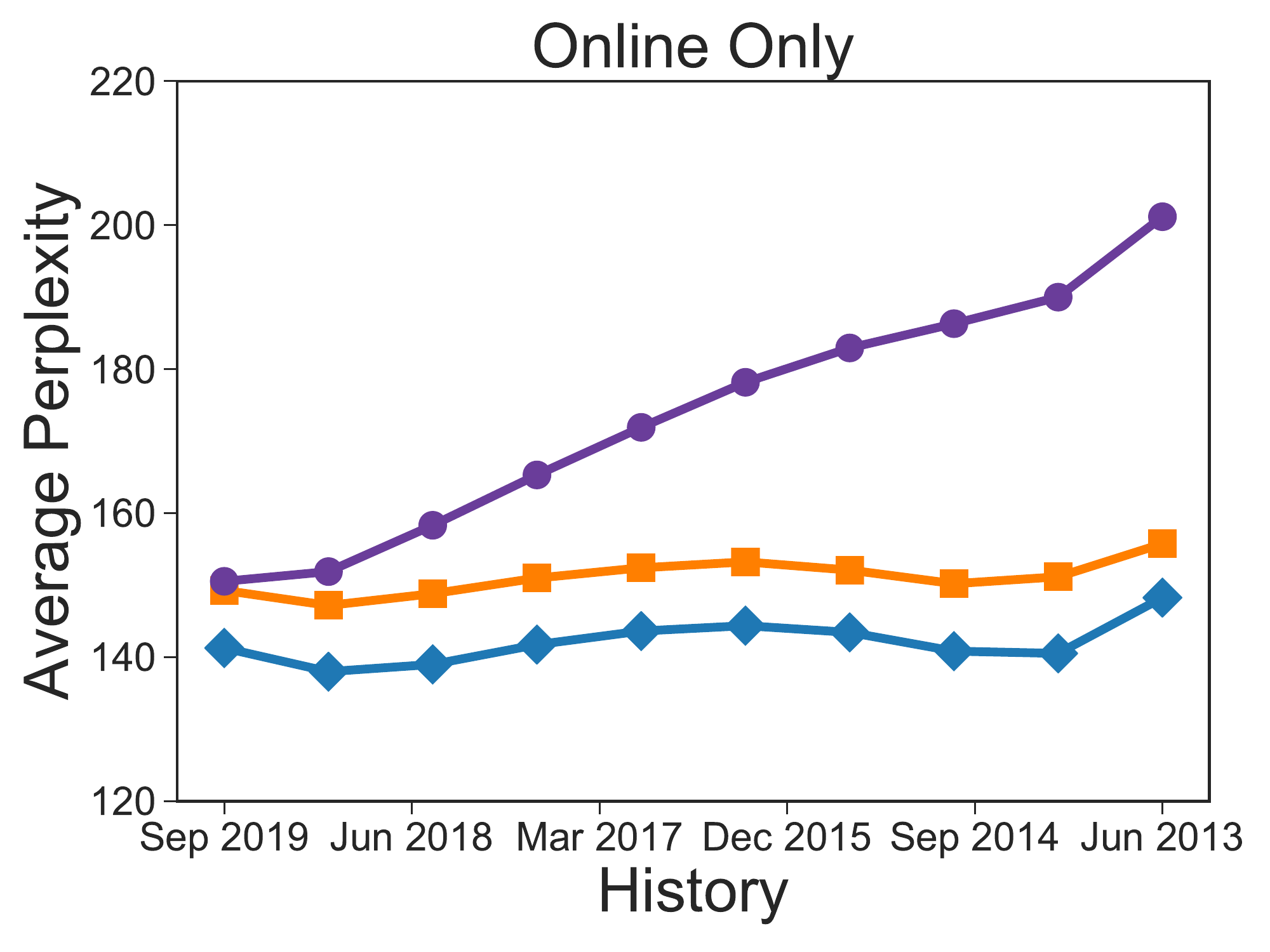} &
        \includegraphics[width=0.245\columnwidth]{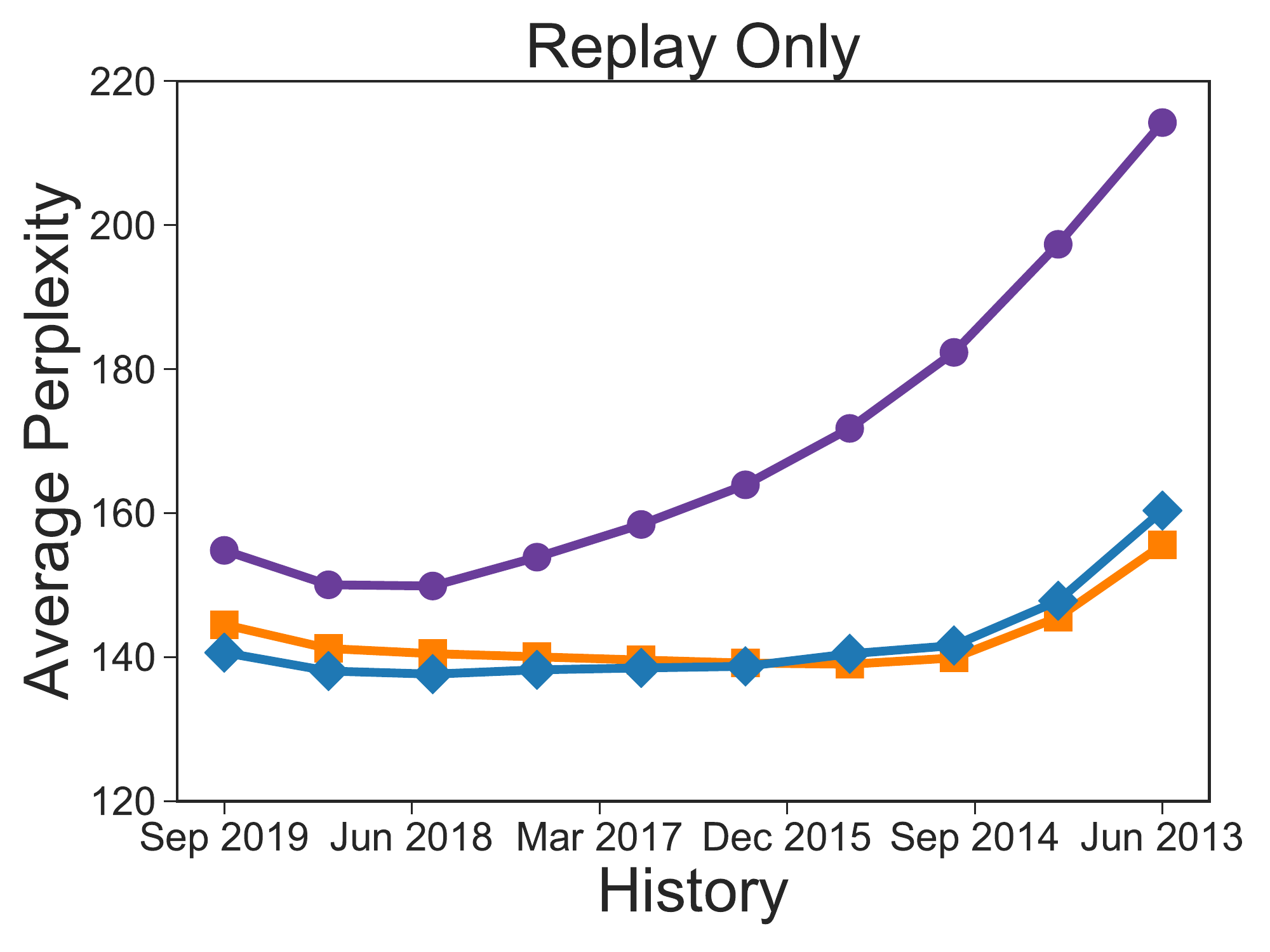} &
        \includegraphics[width=0.245\columnwidth]{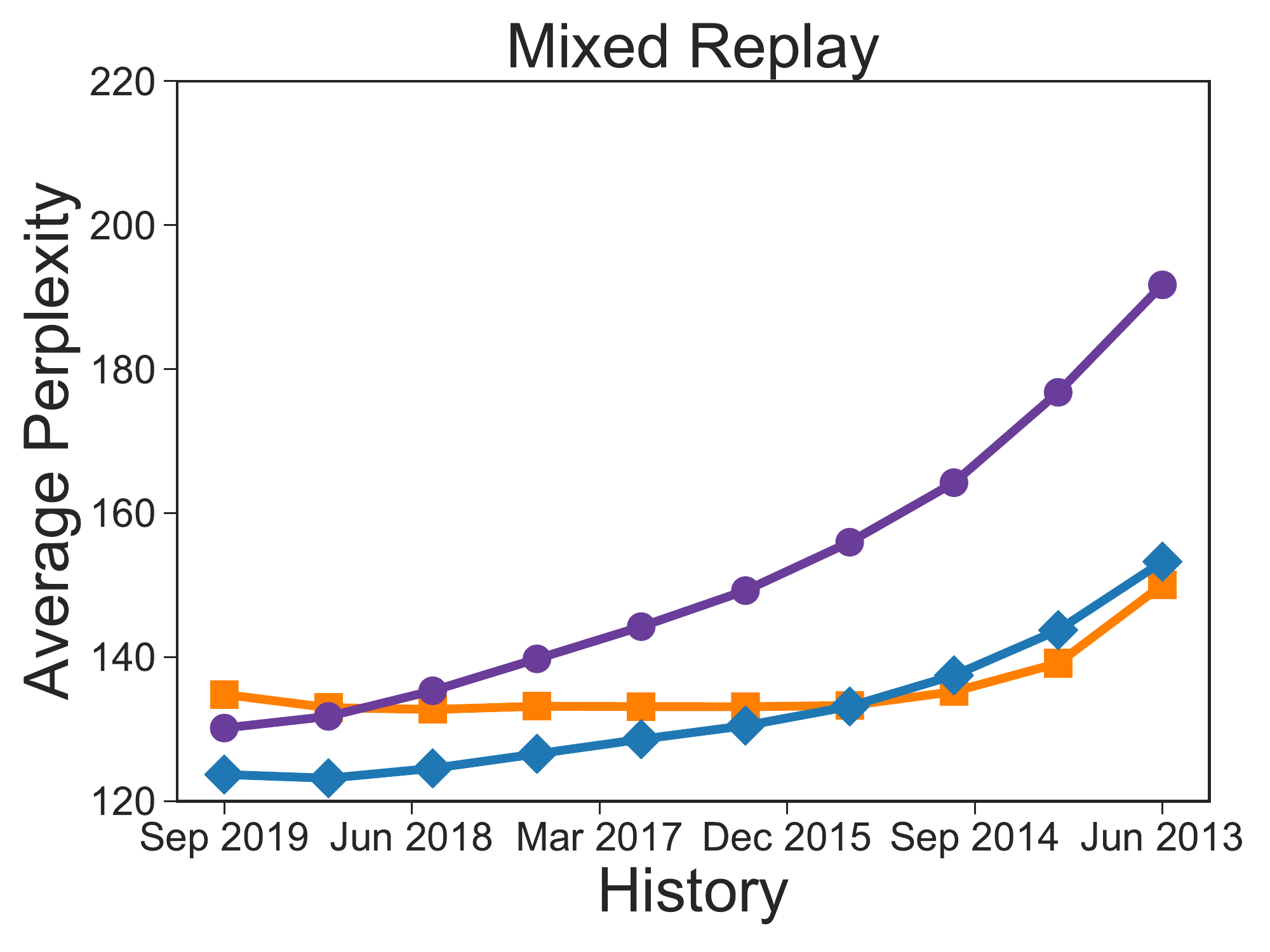}  
    \end{tabular}
    \vspace{-3mm}
    \caption{\textbf{Information retention}. Average perplexity of the final learned model over past data ($\downarrow$).}
    \label{fig:backward}
%    \vspace{-4mm}
\end{figure}

In order to measure forgetting, we compute the perplexity of the final model on historical data for various time differences as $\nicefrac{1}{t}\sum_{s=0}^{t-1}\lL^{\DD_{T-s}}(\btheta_T)$ and plot over $t$ in Figure~\ref{fig:backward}. As expected, all optimizers and models exhibit forgetting (perplexity increases for older data). Interestingly, the optimizer has a significant effect on forgetting, which suggests that the core-set is not chosen well. \ourmethod yields the lowest level of forgetting.

A-GEM retains experience better than other algorithmic models: the difference in perplexity is only 5 points between 2019 and 2013. On the other hand, the perplexity of A-GEM on the latest data is so high that the low level of forgetting is secondary. Other methods yield lower perplexity on historical data despite higher levels of forgetting because they fit the data better. We conclude that the underperformance of A-GEM reported by \citet{chaudhry2019continual} is the fault of poor learning, not the fault of forgetting.

\paragraph{Overall performance.}
A successful continual learner needs to optimize both online learning and information retention. Among algorithmic models, A-GEM forgets the least but also learns the least. \mbox{Online Only} learns the most but also forgets the most. Mixed Replay demonstrates a good trade-off between learning and forgetting. From an optimization perspective, our optimization algorithm, \ourmethod, yields significantly better generalization. \ourmethod and {Online GD} perform similarly in terms of online fit and forgetting. Thus \ourmethod is the superior optimizer overall.

\subsection{Ablation Studies and Analysis}

We perform a number of ablation studies to understand the importance of each proposed component as well as hyperparameter and engineering choices.

\paragraph{Online validation buffer size.}
We evaluate the effect of the validation buffer size in Table~\ref{tab:num_val_size}. We use Mixed Replay $\times$ \ourmethod for this experiment. The results indicate that a small validation buffer (two data points per user, $|\VV| = 180K $) is sufficient to successfully adapt gradient descent. Although one would expect better performance with a bigger validation set, we see lower performance with large buffer sizes. We hypothesize that larger validation buffers impede learning by creating a delay between the time data is observed and the time it is used for learning.

\begin{table}[h]
    \centering
\caption{ \textbf{Validation buffer size.} Word perplexity ($\downarrow$) of Mixed Replay $\times$ \ourmethod.}
\label{tab:num_val_size}
    \begin{tabular}{c@{\hspace{4mm}}c@{\hspace{4mm}}c@{\hspace{4mm}}c@{\hspace{4mm}}c}
        \toprule
         $|\VV| = 90K $ & $|\VV| = 180K $ & $|\VV| = 270K $ & $|\VV| = 360K $ & $|\VV| = 450K $ \\ \midrule
         126.5 & 124.8 & 125.5 & 125.9 & 127.3 \\
        \bottomrule
    \end{tabular}
\end{table}

\begin{figure}[th]
% \begin{wrapfigure}{r}{7cm}
    \centering
%    \vspace{-5mm}
    \includegraphics[width=0.4\columnwidth]{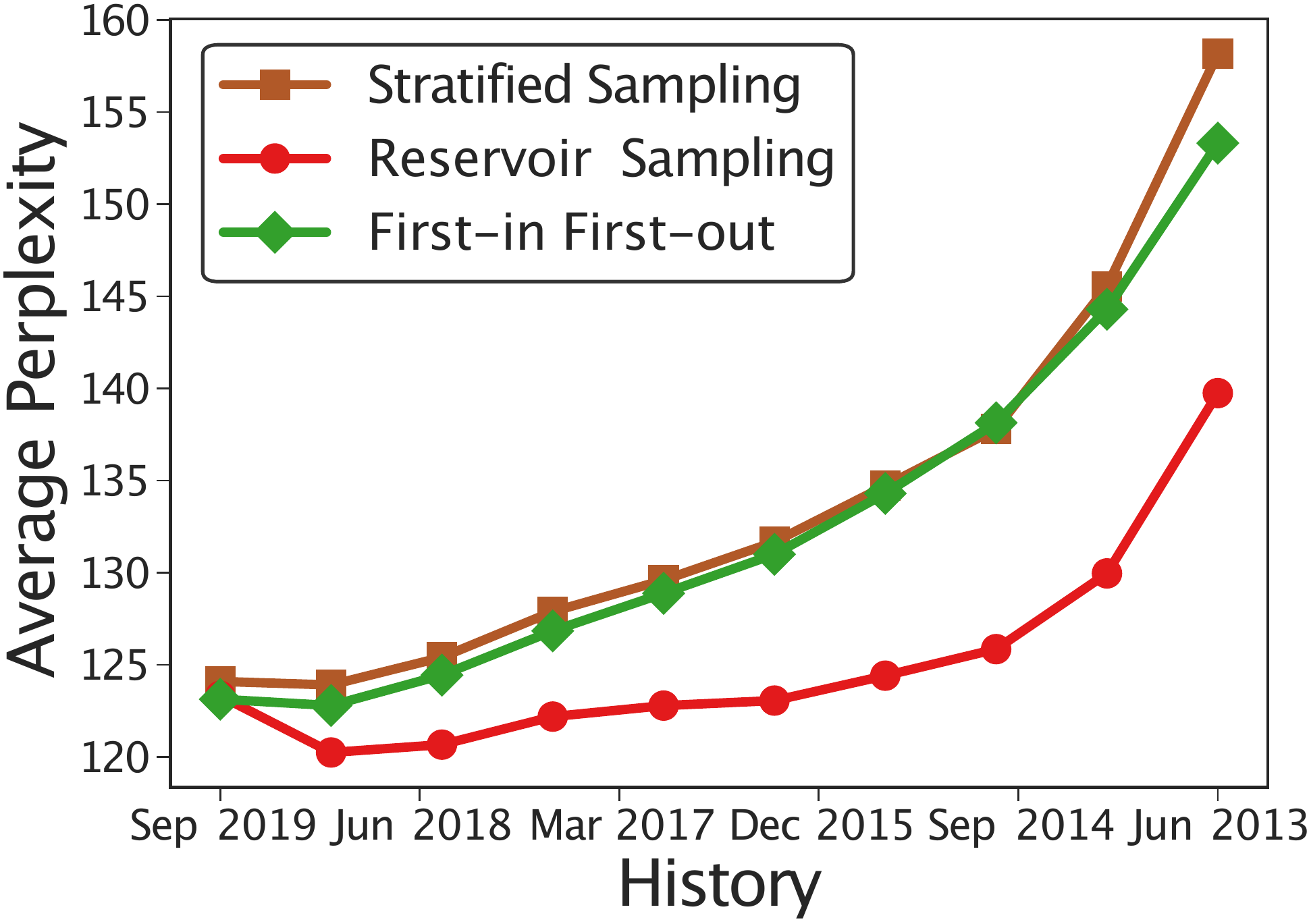}
    \vspace{-4mm}
    \caption{\textbf{Backward transfer} of the final model for \ourdata10M for different validation buffer strategies. We report perplexity on historical data.}
    \label{fig:back_transfer_validation_buffer}
%    \vspace{-10mm}
% \end{wrapfigure}
\end{figure}

\paragraph{Online validation buffer strategy.}
\ourmethod uses a validation buffer to adaptively control the optimization process. A natural validation buffer maintenance scheme is first-in-first-out (\textbf{FIFO}), as it directly approximates the online fit performance. On the other hand, continual learning is a multi-objective problem as it requires simultaneous optimization of test performance, backward transfer, and online fit. If the main objective is test performance, which we define as perplexity over an unseen test set which is balanced over users, a balanced validation buffer would be a better choice. This would result in a \textbf{stratified sampling} scheme where the buffer is kept balanced over users. If the main objective is backward transfer, the intuitive choice becomes \textbf{reservoir sampling} as it ensures the validation buffer is an unbiased estimate of the backward loss. Hence, there is no dominant choice for the validation buffer strategy due to the multi-objective nature of the continual learning problem. We compare these three strategies in terms of test performance in Table~\ref{tab:buffer_strategy}. As expected, the stratified strategy performs best but the difference between methods is minor. We also compare the strategies in terms of backward transfer performance in Figure~\ref{fig:back_transfer_validation_buffer} and conclude that reservoir sampling is the best strategy from a backward transfer perspective. In summary, the choice of validation buffer strategy depends on the partial ordering of the objectives.

\begin{table}[h]
\centering
\caption{ \textbf{Validation buffer strategies.} Word perplexity ($\downarrow$) of Mixed Replay $\times$ \ourmethod.}
\vspace{0.5em}
\label{tab:buffer_strategy}
\begin{tabular}{lccc}
            \toprule
            Buffer strategy & FIFO & Reservoir sampling & Stratified sampling \\ \midrule 
            Perplexity & 127.9  & 127.8 & 126.5 \\ \bottomrule 
\end{tabular}
\end{table}

\paragraph{Effect of $K$.} Table~\ref{tab:num_updates} reports an evaluation of different settings of $K$ for each learner. \ourmethod outperforms the baselines for all $K$. For {Online GD}, $K$ is the number of GD steps. For \ourmethod, $K$ is the maximum number of steps. The results indicate that different models and data call for different settings of $K$.

\begin{table}[H]
    \centering
    \tabcolsep 4pt
    \caption{ \textbf{Effect of $K$}. Word perplexity of test performance ($\downarrow$). \ourmethod outperforms Online GD for all $K$.}
    \vspace{-2mm}
    \begin{tabular}{l@{\hspace{2mm}}c<{\hspace{2mm}}cc<{\hspace{2mm}}cc}
        \toprule
        & & \multicolumn{2}{c<{\hspace{1mm}}}{{Online GD}} & \multicolumn{2}{c}{\ourmethod} \\
        \cmidrule(lr{3mm}){3-4} \cmidrule(l){5-6}
          & {$K = 1$} & {$K = 3$} & {$K = 5$} & {$K = 3$} & {$K = 5$} \\ \midrule
        A-GEM        & $162.1$ & $150.7$ & $149.7$ & $149.0$ & $\mathbf{148.1}$ \\
        Online only  & $153.6$ & $153.1$ & $164.6$ & $\mathbf{145.7}$ & $146.8$ \\
        Replay only  & $142.0$ & $158.1$ & $176.5$ & $\mathbf{139.3}$ & $139.9$ \\
        Mixed replay & $137.2$ & $133.3$ & $139.3$ & $127.0$ & $\mathbf{126.5}$ \\
        \bottomrule
    \end{tabular}
    \label{tab:num_updates}
\end{table}

\subsection{Evaluation on Earlier Coninual Learning Benchmarks} 
We evaluate the \ourmethod optimizer on earlier continual learning benchmarks for the sake of completeness. We follow \citet{aljundi2019gradient} and implement \ourmethod $\times$ \{\text{Online Only}, \text{Replay Only}, \text{Mixed Replay}\} based on the public source code\footnote{Available on https://github.com/rahafaljundi/Gradient-based-Sample-Selection}. For a fair comparison, we used the same hyperparameters as \citet{aljundi2019gradient} to construct the three synthetic datasets (Disjoint CIFAR, Disjoint MNIST, and Permuted MNIST) as well as for creating the baseline algorithms. Following the definition of continual learning, all three datasets are implemented such that only one pass over all the data is available (with randomly permuted data and tasks). We restrict the maximum size of replay memories on each dataset to $300$ and use a two-layer perceptron (with $100$ hidden units) as the learning model. For \ourmethod, we allocate $50$ data points to construct the online validation buffer.

The results are provided in Table~\ref{tab:classic}. \ourmethod consistently improves over continual learning methods that leverage {Online GD}. Even more importantly, we observe that most continual learning algorithms reach similar accuracy, suggesting that traditional benchmarks are not capable of stratifying different algorithms. This further supports the introduction of \ourproblem as a challenging setting for continual learning at scale.

\begin{table}[ht]
%    \vspace{-2mm}
    \tabcolsep 5pt
    \centering
    \caption{\textbf{Evaluation on earlier continual learning benchmarks.} We use final mean accuracy as the metric and report the average over 3 random runs (standard deviation in parentheses). Memory size = 300 for all methods, with one pass over the data as in~\cite{aljundi2019gradient}.}
    \resizebox{\textwidth}{!}{
    \begin{tabular}{@{}l@{\hspace{2mm}}c@{\hspace{2mm}}c@{\hspace{1mm}}c@{\hspace{2mm}}c@{\hspace{1mm}}c@{\hspace{1mm}}c@{\hspace{2mm}}c@{\hspace{1mm}}c@{}}
        \toprule
         & & & \multicolumn{3}{c}{{Online GD}} & \multicolumn{3}{c}{\ourmethod} \\
         \cmidrule(r){4-6} \cmidrule(lr){7-9}
         &GEM &  iCARL&  Online & Replay & Mix. Replay & Online & Replay & Mix. Replay  \\
         \midrule
         Disjoint CIFAR &  $19.1(3.5)$ &  $29.2(5.0)$ & $16.0(0.2)$ & $29.3(0.5)$  & $27.6(2.7)$ & $15.6(0.8)$ & $31.0(0.7)$ & $\bf 31.3(1.2)$ \\
         Disjoint MNIST &  $81.4(1.8)$ &  $84.1(1.2)$ & $19.3(0.1)$ & $85.8(1.4)$  & $82.3(0.3)$ & $29.7(4.2)$ & $\bf 86.2(0.8)$ & $84.2(0.7)$ \\
         Permuted MNIST & $79.6(1.3)$  & $68.2(2.7)$ & $56.9(3.2)$ & $\bf 81.3(0.6)$ & $80.3(1.2)$ & $67.5(0.8)$ & $81.0(0.8)$ & $\bf 81.3(0.2)$ \\ \bottomrule
    \end{tabular}}
    \vspace{1mm}
    \label{tab:classic}
%    \vspace{-6mm}
\end{table}

\section{Conclusion}
We developed personalized online language learning (\ourproblem) as a new setting for continual learning. To support research on this problem, we collected a massive web-scale dataset that comprises 100 million tweets posted over six years (\ourdata). We benchmarked continual learning algorithms on this data and contributed an effective algorithm for continual gradient descent (\ourmethod). Our experiments indicate that there is significant room for progress in continual learning with real web-scale data.

% Acknowledgements should go at the end, before appendices and references

%\acks{Ack.}

% Manual newpage inserted to improve layout of sample file - not
% needed in general before appendices/bibliography.

%\newpage

\bibliography{main}

%\appendix

\end{document}